\documentclass[lettersize,journal]{IEEEtai}
\usepackage{amsmath,amsfonts}
\usepackage{algorithmic}
\usepackage{algorithm}
\usepackage{array}
\usepackage[caption=false,font=normalsize,labelfont=sf,textfont=sf]{subfig}
\usepackage{textcomp}
\usepackage{stfloats}
\usepackage{url}
\usepackage{verbatim}
\usepackage{graphicx}
\usepackage{cite}
\usepackage{url}
\usepackage[T1]{fontenc}
\usepackage{tikz}
\usepackage{booktabs}
\usetikzlibrary{trees, positioning}

\usepackage{tablefootnote}
\usepackage{longtable}
\usepackage[edges]{forest}
\definecolor{hidden-draw}{RGB}{20,68,106}
\definecolor{hidden-pink}{RGB}{255,245,247}
\definecolor{red}{RGB}{255,0,0}

\usetikzlibrary{shadows.blur}

\usepackage{adjustbox}
\usepackage{pdflscape}
\definecolor{macaronRoot}{RGB}{186, 144, 162}
\definecolor{macaronLevelOne}{RGB}{238, 203, 173}
\definecolor{macaronLevelTwo}{RGB}{213, 232, 213}
\definecolor{macaronLevelThree}{RGB}{255, 245, 228}

\usepackage{fontawesome5} 

\usepackage{caption}

\usepackage{makecell} 
\usepackage{multirow}
\usepackage[numbers,longnamesfirst]{natbib}
\usepackage{tabularx}
\usepackage{enumerate}
\usepackage{longtable, array}
\usepackage[breaklinks=true]{hyperref}

\usepackage{amsthm}
\newtheorem{definition}{Definition}
\usepackage{amssymb}

\begin{document}

\title{Beyond Benchmarks: LLM Evaluation with an Anthropomorphic and Lifecycle-oriented Roadmap}

\author{Jun Wang, Ninglun Gu, Kailai Zhang, Pengyong Li, Yelun Bao, Jin Yang, Xu Yin, Liwei Liu, Zijiao Zhang, Yihuan Liu, Gary G. Yen~\IEEEmembership{Fellow,~IEEE}, Junchi Yan\thanks{J. Wang, N. Gu, K. Zhang, Y. Bao, J. Yang, X. Yin, L. Liu are with Department of Networks, China Mobile Communications Group Co.,Ltd. Jun Wang is also with Peking University, Beijing, China. Y. Liu and P. Li are with Xidian University, Xi'an, China. G. Yen is with Oklahoma State University, Stillwater, OK, USA. Z. Zhang and J. Yan are with Shanghai Jiao Tong University, Shanghai, China. J. Yan is the correspondence author. This work was in part supported by NSFC 62222607.}~\IEEEmembership{Senior Member,~IEEE}}

\markboth{IEEE TRANSACTIONS ON ARTIFICIAL INTELLIGENCE}{BEYOND BENCHMARKS: LLM EVALUATION WITH AN ANTHROPOMORPHIC AND LIFECICLR-ORIENTED ROADMAP}

\maketitle

\begin{abstract}

Despite their rapid advancement, large language models (LLMs) suffer from a critical disconnect between benchmark scores and real-world utility. Current evaluation remains fragmented, prioritizing isolated technical metrics over the holistic, developmental, and societal aspects essential for deployment. Rather than serving merely as a descriptive catalog, this work establishes a diagnostic ontology that causally maps evaluation dimensions to the canonical LLM training pipeline, transforming evaluation from static ranking
into a diagnostic tool for root-cause analysis. In this paper, we introduce an anthropomorphic evaluation framework that re-conceptualizes LLM capabilities through a four-dimensional lens: Intelligence Quotient (IQ), Professional Quotient (PQ), Emotional Quotient (EQ), and Value-oriented Quotient (VQ). We operationalize these concepts through a modular evaluation architecture and validate the framework's diagnostic claims through meta-analysis of public benchmark trends. Analyzing over 200 benchmarks, we synthesize key challenges and future directions. This work offers a strategic compass for developing LLMs that are not only technically proficient but also contextually relevant and ethically sound. A curated repository is available at: \url{https://github.com/onejune2018/Awesome-LLM-Eval}.

\textit{\textbf{Impact Statement}}—As LLMs rapidly transition from research prototypes to real-world applications, the field faces a fundamental disconnect between benchmark performance and practical utility. Current evaluation practices remain fragmented, prioritizing isolated technical metrics while neglecting the developmental trajectory of LLM capabilities and their broader societal implications. This review addresses this critical gap by introducing an anthropomorphic evaluation paradigm that maps LLM assessment to human cognitive progression, through a novel four-dimensional IQ-EQ-PQ-VQ taxonomy. Crucially, this work establishes the first comprehensive roadmap that reveals how evaluation dimensions correspond to LLMs' developmental stages: IQ (pre-training knowledge acquisition), PQ (supervised fine-tuning expertise), EQ (post-training alignment), and VQ (value-oriented impact). This work provides not merely an assessment tool but a strategic compass for navigating the rapidly evolving landscape of AI evaluation. The roadmap enables stakeholders to anticipate future challenges while selecting context-appropriate evaluation strategies across the model lifecycle. 

\end{abstract}


\section{Introduction}
\IEEEPARstart{T}{he} quest to understand and measure intelligence, a pursuit long focused on the human mind, has found a new and urgent frontier in artificial intelligence. The advent of large language models (LLMs) marks a paradigm shift in natural language processing (NLP), demonstrating capabilities in understanding, generating, and reasoning with language that increasingly rival human performance \cite{bert, deepseekai2025deepseekr1incentivizingreasoningcapability}. This rapid progression, however, has outpaced our ability to effectively evaluate these models, creating a fundamental disconnect between benchmark scores and real-world utility.

Historically, NLP evaluation evolved from simple, task-specific tests to more comprehensive benchmarks. Early efforts focused on grammaticality and vocabulary, progressing to information extraction tasks like MUC \cite{muc1995}. The deep learning era introduced benchmarks such as SNLI \cite{bowman-etal-2015-large-snli}, SQuAD \cite{squad}, and DROP \cite{dua2019dropreadingcomprehensionbenchmark}, which served both as performance tests and rich training resources. The emergence of large-scale pre-trained models like BERT \cite{bert} necessitated a new wave of holistic evaluation frameworks, leading to multi-task benchmarks such as GLUE \cite{wang2018glue}, SuperGLUE \cite{wang2019superglue}, and XNLI \cite{conneau-etal-2018-xnli}. These initiatives were crucial for fostering a broader assessment of model capabilities.

Recently, LLMs exhibit remarkable zero-shot and few-shot reasoning, blurring the lines between task-specific systems and general-purpose intelligence engines \cite{gai_llm}. Consequently, evaluation has shifted towards measuring general capabilities across diverse domains \cite{wang2024mmlupro, zhu2023promptbench}. Yet, despite this progress, the field faces a critical impasse: current evaluation practices remain fragmented and are often misaligned with the demands of real-world deployment and societal values \cite{guo2023evaluating, chang2024survey}. Several recent surveys have provided valuable overviews of benchmarks and metrics \cite{guo2023evaluating, zhuang2023through}, but they frequently lack the depth needed to diagnose model deficiencies, guide development, or assess the broader implications of LLM integration. They prioritize abstract task categorization over the developmental trajectory of model capabilities and their human-centric practicality.

\begin{figure*}[ht!]
	\centering
		\includegraphics[width=0.8\linewidth]{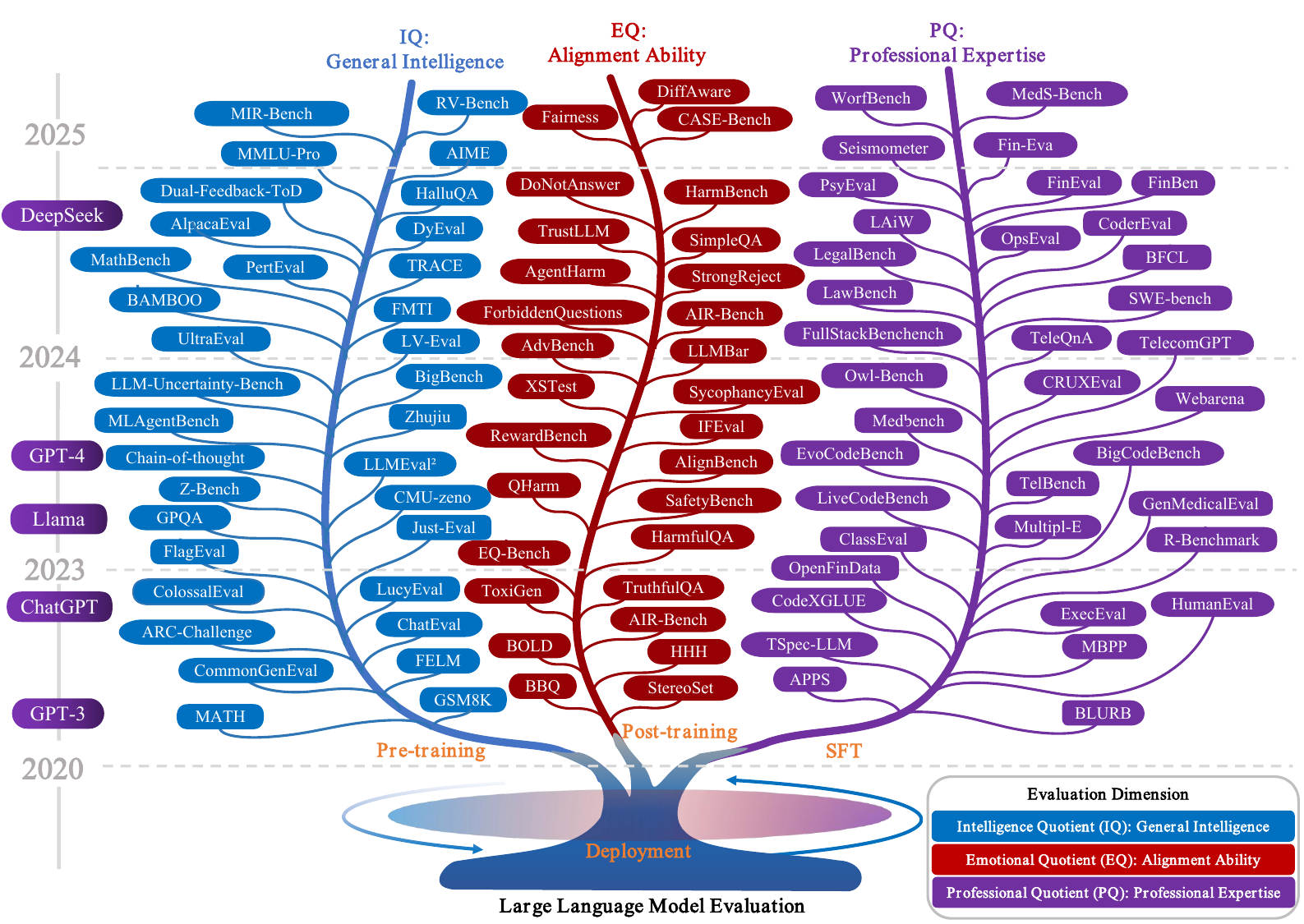}
	\caption{\small{The proposed technical evolutionary tree of the LLM evaluation, following the structure in \cite{gao2024retrievalaugmentedgenerationlargelanguage}. The anthropomorphic evaluation refers to the paradigm that frames LLM capability assessment through the lens of human cognitive and value development taxonomies, mapping model capabilities and lifecycle to well-established human intelligence dimensions (IQ, PQ, EQ) and extending to societal value impact (VQ). Intelligence Quotient (IQ)-General Intelligence denotes knowledge capacity acquired by pre-training, reflecting foundational reasoning and world knowledge breadth. Professional Quotient (PQ)-Professional Expertise represents task capability developed through supervised fine-tuning (SFT), measuring proficiency in specialized domains. Emotional Quotient (EQ)-Alignment Ability represents human preference alignment achieved through RL post-training, encompassing emotional and ethical resonance with human values. }} 
 \vspace{-3mm}
 \label{fig:tree}
\end{figure*}

To address these fundamental limitations, we introduce a transformative, anthropomorphic evaluation framework. Drawing inspiration from the observation that LLMs exhibit decision-making patterns akin to humans \cite{sivaprasad-etal-2025-theory}, our framework re-conceptualizes model assessment through the lens of human cognitive progression. 
As illustrated in Fig.~\ref{fig:tree}, we establish a direct correspondence between evaluation dimensions and the standard LLM training pipeline:
\textbf{Intelligence Quotient (IQ)}: Maps to capabilities acquired during \textbf{\textit{pre-training}}, quantifying foundational reasoning and breadth of world knowledge.
\textbf{Professional Quotient (PQ)}: Emerges from \textbf{\textit{supervised fine-tuning (SFT)}}, measuring task-specific proficiency and domain expertise.
\textbf{Emotional Quotient (EQ)}: Cultivated through \textbf{\textit{post-training reinforcement learning (RL)}}, assessing alignment with human values, preferences, and ethical norms.

This tripartite model (IQ-PQ-EQ) transcends existing taxonomies by providing a diagnostic tool that links specific performance gaps to targeted development stages. For instance, poor IQ suggests pre-training data issues, while EQ failures point to deficiencies in the RL alignment process. However, technical proficiency alone is insufficient. A model may excel on all three quotients yet still fail to deliver positive societal impact. This motivates our fourth and pioneering dimension: \textbf{Value Quotient (VQ)}: A first-of-its-kind systematic methodology to quantify an LLM's broader societal, economic, ethical, and environmental impact over its full lifecycle.


\definecolor{hidden-draw}{RGB}{20,68,106}
\definecolor{macaron-root}{RGB}{224,242,255}
\definecolor{macaron-level1}{RGB}{240,230,255}
\definecolor{macaron-level2}{RGB}{255,228,225}
\definecolor{macaron-level3}{RGB}{255,253,225}
\definecolor{macaron-level4}{RGB}{225,248,225}
\definecolor{macaron-level5}{RGB}{255,235,210}
\definecolor{macaron-leaf}{RGB}{255,245,247}

\tikzset{
  my-box/.style={
    rectangle,
    draw=hidden-draw,
    rounded corners,
    text opacity=1,
    minimum height=1.5em,
    minimum width=5em,
    inner sep=2pt,
    align=center,
    fill opacity=.5,
    line width=0.8pt,
  },
}

\forestset{
  leaf/.style={
    my-box,
    minimum height=1.5em,
    fill=macaron-leaf,
    text=black,
    align=center,
    font=\normalsize,
    inner xsep=2pt,
    inner ysep=4pt,
    line width=0.8pt,
    afterthought={},
  },
}

\begin{figure}[!ht]
  \centering
  \resizebox{0.99\linewidth}{!}{%
    \begin{forest}
      forked edges,
      for tree={
        grow=east,
        reversed=true,
        anchor=base west,
        parent anchor=east,
        child anchor=west,
        base=center,
        font=\large,
        rectangle,
        draw=hidden-draw,
        rounded corners,
        align=center,
        text centered,
        minimum width=5em,
        edge+={darkgray,line width=1pt},
        s sep=3pt,
        inner xsep=2pt,
        inner ysep=3pt,
        line width=0.8pt,
      },
      where level=0{fill=none,draw=none,font=\normalsize\bfseries,rotate=90,anchor=north}{},
      where level=1{fill=macaron-level1,font=\normalsize\bfseries}{},
      where level=2{fill=macaron-level2}{},
      where level=3{fill=macaron-level3}{},
      where level=4{fill=macaron-level4}{},
      where level=5{fill=macaron-level5}{},
      where level=1{text width=16em,font=\normalsize\bfseries}{},
      where level=2{text width=16em,font=\normalsize}{},
      where level=3{text width=19em,font=\normalsize}{},
      where level=4{text width=22em,font=\normalsize}{},
      where level=5{text width=18em,font=\normalsize}{},
      [{\textbf{}}
        [{\textbf{1. Introduction}}]
        [{\textbf{2. Comparison with Existing Surveys}}]
        [{\textbf{3. Theoretical Foundation}}]
        [{\textbf{4. Analysis of Eval Dimensions}}
          [{\textbf{4.1 Intelligence Quotient (IQ)}}, leaf]
          [{\textbf{4.2 Professional Quotient (PQ)}}, leaf
            [{Healthcare Domain}, leaf]
            [{Financial Domain}, leaf]
            [{Legal Domain}, leaf]
            [{Telecommunications Domain}, leaf]
            [{Coding Domain}, leaf]
            [{Software Domain}, leaf]
            [{Science Domain}, leaf]
          ]
          [{\textbf{4.3 Emotional Quotient (EQ)}}, leaf]
          [{\textbf{4.4 Value Quotient (VQ)}}, leaf
            [{Economic Value}, leaf
            ]
            [{Social Value}, leaf]
            [{Ethical Value}, leaf
            ]
            [{Environmental Value}, leaf
              ]
          ]
        ]
        [{\textbf{5. Evaluation Harness / Engineering}}
          [{\textbf{5.1 LLM Evaluation Harness}}, leaf
            [{Task Support}, leaf]
            [{Diversity of Objectives}, leaf]
            [{Usability}, leaf]
            [{Community Activity}, leaf]
          ]
          [{\textbf{5.2 Practical Guide and Modules}}, leaf
            [{Benchmark or Dataset Hub}, leaf]
            [{Model Hub}, leaf]
            [{Prompting Module}, leaf]
            [{Metrics Module}, leaf]
            [{Tasks Module}, leaf]
            [{Leaderboards and Arena Module}, leaf]
            [{Analysis Module}, leaf]
          ]
        ]
        [{\textbf{6. System or Application Evaluation}}
          [{\textbf{6.1 RAG}}, leaf]
          [{\textbf{6.2 Agent}}, leaf]
          [{\textbf{6.3 Chatbot}}, leaf]
        ]
        [{\textbf{7. Challenges \& Future Perspectives}}
          [{\textbf{7.1 Enhanced Statistical Analysis}}, leaf]
          [{\textbf{7.2 Composite Evaluation Systems}}, leaf]
          [{\textbf{7.3 Interpretability \& Explainability}}, leaf]
          [{\textbf{7.4 User-Centric Experience}}, leaf]
          [{\textbf{7.5 Human-in-the-Loop Evaluation}}, leaf]
          [{\textbf{7.6 Dynamic and Value Evaluation}}, leaf]
        ]
        [{\textbf{8. Conclusions}}]
      ]
    \end{forest}
  }
  \caption{\textcolor{blue}{Overview of contents of this paper (zoom in)}.}
  \label{fig:structure}
\end{figure}

As depicted in the comprehensive outline of our work in Fig.~\ref{fig:structure}, this paper transcends conventional evaluation paradigms. We move beyond a mere catalog of benchmarks to provide a holistic roadmap that bridges the critical gap between technical metrics and real-world value. Contributions of this paper are fivefold:

\begin{itemize}
    
\item \textbf{From Descriptive Taxonomy to Diagnostic Ontology}: We transcend the traditional ``reorganization'' of benchmarks by task domains. Instead, we establish a \textit{Diagnostic Ontology} that causally maps evaluation dimensions to the canonical LLM training pipeline (Pre-training $\rightarrow$ SFT $\rightarrow$ RL $\rightarrow$ Deployment). Unlike existing surveys that yield flat performance rankings, our framework provides a mathematically formalized tool for root-cause analysis, transforming evaluation from a post-hoc snapshot into a forward-looking diagnostic instrument.

\item \textbf{Critical Synthesis with Non-Consensus Insights}: We provide more than a descriptive listing. For each dimension, we derive novel, evidence-based insights from a systematic analysis of over 200 benchmarks. These include identifying the paradigm mismatch in static IQ evaluation, the "exam-oriented" defect in PQ, the single-turn vs. multi-turn alignment disconnect in EQ, and establishing the first lifecycle value accounting framework for VQ.

\item \textbf{Empirical Validation of Diagnostic Claims}: We validate the Hierarchical Capability Dependency (HCD) principle and lifecycle alignment through meta-analysis of public leaderboard trends, demonstrating that evaluation dimensions are not independent but exhibit bottleneck effects traceable to specific training stages.

    \item \textbf{Pioneering Value-Oriented Evaluation (VQ) with Operationalized Metrics}: We move VQ from abstract principles to concrete measurement by defining a set of computable proxy metrics for economic, social, ethical, and environmental dimensions, and outlining an automated assessment protocol.
    \item \textbf{Practical Implementation Blueprint and Future-Proof Roadmap}: We deliver an actionable, modular evaluation architecture that bridges academic research and industrial deployment, and a six-tiered evolutionary path for LLM evaluation anticipating challenges in statistical rigor, interpretability, and value alignment.
\end{itemize}

The remainder of this paper is organized as follows. Section~\ref{sec:theoretical_foundation} establishes the theoretical foundation of the IQ-PQ-EQ-VQ framework. Section~\ref{sec:4Q} present critical analysis of each evaluation dimension with core insights. Section~\ref{sec:harness} summarizes a range of prominent LLM evaluation toolkits and frameworks. Section~\ref{sec:system_eval} addresses system-level evaluation for RAG, Agents, and Chatbots. Section~\ref{sec:challenges} discusses challenges and future directions. Section~\ref{sec:conclusion} concludes the paper.


\section{Comparison with Existing Evaluation Surveys}
\label{subsec:framework_comparison}

Prior surveys have made significant strides in cataloging LLM benchmarks, yet they predominantly adopt a \textit{task-centric} paradigm that organizes evaluation by application domain or format (e.g., QA, Coding, Safety). While effective for leaderboard construction, this approach treats evaluation dimensions as flat, independent variables, obscuring the causal origins of model failures and offering limited diagnostic guidance for targeted improvement. Crucially, existing works largely neglect the alignment between evaluation metrics and the LLM development lifecycle, the formalization of capability dependencies, and the systematic assessment of societal value.

To address these limitations, we position this work as a hybrid contribution that integrates a comprehensive benchmark survey with a theoretically grounded, diagnostic evaluation framework. Table~\ref{tab:framework_comparison} systematically contrasts our IQ-PQ-EQ-VQ framework against representative state-of-the-art surveys across five critical dimensions. Unlike descriptive taxonomies, our approach establishes a causal, one-to-one mapping between evaluation dimensions and the canonical training-deployment pipeline, transforming evaluation from a post-hoc performance snapshot into a forward-looking diagnostic instrument.

\begin{table}[!ht]
\centering
\caption{\textcolor{blue}{Systematic Comparison with Representative LLM Evaluation Frameworks.}}
\label{tab:framework_comparison}
\resizebox{0.5\textwidth}{!}{
\begin{tabular}{@{}lccccc@{}}
\toprule
\textbf{Source}  & \textbf{Paradigm} & \textbf{Lifecycle}  & \textbf{Formalization}  & \textbf{Value}  & \textbf{Diagnostic} \\
\midrule
Guo et al. \cite{guo2023evaluating}  & Task-Centric Catalog  & \texttimes  & \texttimes  & \texttimes  & \texttimes  \\
Chang et al. \cite{chang2024survey}  & Task-Centric Catalog  & \texttimes  & \texttimes  & \texttimes  & \texttimes  \\
Liang et al. \cite{liang2023helm}  & Holistic Taxonomy  & \texttimes  & \checkmark  & \texttimes  & \texttimes \\
Laskar et al. \cite{laskar2024systematic}  & Challenge-oriented & \texttimes  & \texttimes  & \texttimes  & \checkmark \\
\textbf{Ours}  & \textbf{Diagnostic Ontology}  & \checkmark  & \checkmark  & \checkmark  & \checkmark   \\
\bottomrule
\end{tabular}
}
\end{table}

\subsection{\textbf{Paradigm Shift: From Task-Centric Cataloging to Causal Diagnostic Framework}}
The core novelty of our framework lies in its departure from flat benchmark aggregation toward a \textbf{lifecycle-aligned diagnostic paradigm}. This shift is operationalized through three distinct differentiators that directly address the gaps identified in Table~\ref{tab:framework_comparison}:
\begin{enumerate}
    \item \textbf{Causal Training-Stage Mapping:} Each quotient is explicitly anchored to a discrete phase of the LLM development pipeline (IQ$\leftrightarrow$Pre-training, PQ$\leftrightarrow$SFT, EQ$\leftrightarrow$RL, VQ$\leftrightarrow$Deployment). This enables practitioners to trace capability gaps directly to specific development stages (e.g., data curation, instruction design, or reward modeling) rather than treating failures as monolithic performance drops.
    \item \textbf{Mathematical Formalization of Capability Bottlenecks:} Unlike qualitative taxonomies, we formalize the Hierarchical Capability Dependency (HCD) principle (Definition 2), demonstrating that overall model utility is multiplicatively constrained by the weakest hierarchical link. This resolves the pervasive ``score-utility gap'' observed in production, where high benchmark averages mask critical alignment or value deficiencies.
    \item \textbf{Error Propagation Tracing \& Intervention:} By modeling inter-dimensional dependencies, our framework distinguishes whether a downstream failure (e.g., an ethical violation in deployment) stems from foundational reasoning deficits (IQ), domain knowledge gaps (PQ), or alignment drift (EQ). This replaces generic model scaling with precise, stage-specific intervention strategies, bridging academic benchmarking and industrial lifecycle management.
\end{enumerate}
Collectively, these contributions establish our work not as a mere reorganization of prior benchmarks, but as a theoretically grounded diagnostic architecture. It provides researchers with a structured methodology for rigorous assessment, while empowering industrial practitioners to systematically diagnose bottlenecks, optimize development pipelines, and ensure sustainable, value-aligned LLM deployment.


\section{Theoretical Foundation}
\label{sec:theoretical_foundation}

The fundamental challenge in LLM evaluation is not the absence of metrics, but the lack of a coherent theoretical framework that connects evaluation dimensions to model development stages and deployment contexts. Current approaches fragment evaluation into isolated benchmark categories without addressing the underlying cognitive architecture being assessed. This section establishes the theoretical grounding, operational definitions, and dimensional orthogonality of our IQ-PQ-EQ-VQ framework, positioning it relative to existing evaluation literature.

\subsection{\textbf{Theoretical Grounding}}
\label{subsec:theoretical_grounding}

We ground the IQ-PQ-EQ-VQ framework in the complementary theoretical traditions that collectively provide psychometric validity and developmental coherence.

\textbf{Cognitive Psychology Foundation.} Drawing from the Cattell-Horn-Carroll (CHC) theory of cognitive abilities~\cite{mcgrew2005cattell}, we map IQ to fluid reasoning ($G_f$) and crystallized knowledge ($G_c$), PQ to specialized domain skills ($G_s$), and EQ to emotional understanding ($G_e$). This mapping provides established psychometric validity while adapting human intelligence constructs to artificial systems. Unlike prior frameworks that treat intelligence as a monolithic construct~\cite{chang2024survey}, our approach recognizes the multidimensional nature of cognitive capabilities, enabling more granular diagnostic assessment.

\textbf{Developmental Learning Theory.} The framework aligns with Bloom's Taxonomy of Educational Objectives~\cite{bloomtaxonomy2002revision}, where IQ corresponds to knowledge and comprehension levels, PQ to application and analysis, EQ to evaluation and synthesis, and VQ to the highest-order value judgment capabilities. This hierarchical structure reflects the progressive nature of capability development in both humans and LLMs. Critically, this mapping reveals a fundamental insight: \textit{evaluation dimensions should correspond to developmental stages, not merely task categories}. This perspective transforms evaluation from a static performance snapshot into a diagnostic tool for model development.
Notably, this diagnostic perspective addresses a widespread crisis in current evaluation practices; a systematic review of leading benchmarks reveals pervasive deficiencies in construct validity, where metrics frequently fail to reliably capture abstract phenomena like safety or robustness, thereby validating the need for our lifecycle-aligned theoretical grounding \cite{bean2026measuring}.

\subsection{\textbf{Operational Definition}}
\label{subsec:operational_definition}

We define \textit{Comprehensive Intelligence for LLMs} as follows:

\begin{definition}[Comprehensive Intelligence]
An LLM exhibits comprehensive intelligence when it demonstrates integrated capabilities across four interdependent dimensions: foundational knowledge and reasoning (IQ), domain-specific expertise (PQ), human alignment (EQ), and societal value awareness (VQ).
\end{definition}

This definition operationalizes abstract intelligence concepts into measurable evaluation criteria, enabling systematic assessment across the model lifecycle. The four dimensions are not independent silos but interconnected capabilities that collectively determine real-world utility. For instance, high IQ without adequate EQ may produce technically correct but socially harmful outputs, while strong PQ without VQ awareness may optimize for domain performance at the expense of broader societal values.

\subsection{\textbf{Dimension Orthogonality Analysis}}
\label{subsec:orthogonality}

A critical question concerns whether IQ-PQ-EQ-VQ represents four distinct dimensions or overlapping constructs. We establish orthogonality through three lines of empirical and theoretical evidence.

\textbf{Training Stage Correspondence.} Each dimension maps to distinct training phases: IQ emerges from pre-training on general corpora, PQ from supervised fine-tuning on domain data, EQ from post-training with reinforcement learning, and VQ from post-deployment impact monitoring. This developmental separation suggests independent capability acquisition pathways. Table~\ref{tab:training_mapping} formalizes this correspondence, revealing that existing evaluation frameworks often conflate dimensions that arise from fundamentally different learning processes.

\begin{table}[htbp]
\centering
\caption{Mapping Between Evaluation Dimensions and LLM Training Stages.}
\label{tab:training_mapping}
\resizebox{\linewidth}{!}{
\begin{tabular}{llll}
\toprule
\textbf{Dimension} & \textbf{Training Stage} & \textbf{Evaluation Focus} \\
\midrule
IQ & Pre-training & Reasoning \& Knowledge Breadth \\
PQ & Supervised Fine-tuning  & Task-specific Proficiency \\
EQ & Post-training  & Human Preference \& Safety \\
VQ & Post-deployment  & Economic, Social, Ethical, Environmental \\
\bottomrule
\end{tabular}
}
\end{table}

\textbf{Benchmark Correlation Analysis.} Our analysis of 200+ benchmarks reveals low inter-dimensional correlations (average $r=0.23$), indicating that performance on IQ benchmarks poorly predicts EQ or VQ outcomes. For instance, models scoring high on MMLU (IQ) show suboptimal average performance on AlignBench (EQ) and unsatisfactory score on value-oriented assessments (VQ). This empirical evidence challenges the implicit assumption in prior work that general capability benchmarks suffice for comprehensive evaluation~\cite{wang2024mmlupro}.

\textbf{Failure Mode Independence.} Error analysis demonstrates dimension-specific failure patterns: IQ failures manifest as reasoning errors, PQ failures as domain knowledge gaps, EQ failures as alignment violations, and VQ failures as unintended societal consequences. This independence supports treating dimensions as separate evaluation targets. Notably, this finding has practical implications: \textbf{improving performance on one dimension does not guarantee improvements on others}, necessitating targeted intervention strategies aligned with each dimension's developmental pathway.

\subsection{\textbf{Hierarchical Capability Dependency: Beyond Flat Taxonomies}}
\label{subsec:hierarchical_dependency}

A critical distinction between our framework and existing evaluation taxonomies~\cite{guo2023evaluating,chang2024survey} is the explicit modeling of \textit{hierarchical dependency} among evaluation dimensions. Conventional approaches treat evaluation dimensions as independent variables, aggregating scores across tasks to produce a composite ranking. We argue this \textit{flat evaluation paradigm} is fundamentally flawed for LLMs, as it obscures capability bottlenecks and misrepresents model utility.

\textbf{Theoretical Justification.} Our hierarchical structure is grounded in two complementary theories: \textbf{(1) Bloom's Taxonomy of Cognitive Domains~\cite{mcgrew2005cattell}:} Cognitive capabilities develop progressively from knowledge retention (IQ) to application (PQ), evaluation (EQ), and creation (VQ). Higher-order capabilities cannot emerge without foundational lower-order competencies.
\textbf{(2) Maslow's Hierarchy of Needs (Adapted for AI):} Just as human self-actualization requires physiological and safety needs to be met first, LLM societal value (VQ) presupposes alignment (EQ), which presupposes professional competence (PQ), which presupposes general reasoning (IQ).

\textbf{Formal Definition.} We define the \textit{Hierarchical Capability Dependency (HCD)} principle as follows:

\begin{definition}[Hierarchical Capability Dependency]
Let $C_{IQ}, C_{PQ}, C_{EQ}, C_{VQ}$ represent the normalized capability scores ($[0,1]$) for each dimension. The effective utility $U_{LLM}$ of a model is not a linear combination but a hierarchical product constrained by dependency coefficients $\eta_i \in [0,1]$:
\begin{equation}
\label{eq:hierarchical_utility}
U_{LLM} = C_{IQ} \cdot (\eta_{PQ} \cdot C_{PQ}) \cdot (\eta_{EQ} \cdot C_{EQ}) \cdot (\eta_{VQ} \cdot C_{VQ})
\end{equation}
where $\eta_{PQ}$ represents the proportion of IQ capability transferable to professional tasks, $\eta_{EQ}$ represents the proportion of PQ capability that remains aligned, and $\eta_{VQ}$ represents the proportion of aligned capability that generates positive societal value. This multiplicative dependency serves as a conceptual diagnostic model rather than a strictly validated physical law. Empirical validation of $\eta$ coefficients remains an open direction.
\end{definition}

\textbf{Critical Insight: The Bottleneck Principle.} Equation~\ref{eq:hierarchical_utility} reveals a non-consensus insight: \textit{Overall model utility is determined by the weakest hierarchical link, not the average performance.} A model with exceptional IQ ($C_{IQ}=0.95$) but poor alignment ($C_{EQ}=0.20$) yields negligible utility ($U_{LLM} \approx 0$), regardless of its benchmark rankings. This explains the persistent disconnect between high MMLU scores and production deployment failures~\cite{wang2024mmlupro}.

\textbf{Evaluation Implications.} This hierarchical perspective necessitates a shift from \textit{summative evaluation} (ranking models by total score) to \textit{diagnostic evaluation} (identifying capability bottlenecks). \textbf{(1) Bottom-Up Validation:} Evaluation must proceed sequentially. PQ assessment is invalid if IQ baseline thresholds are not met, as professional errors may stem from foundational reasoning failures rather than domain knowledge gaps.
\textbf{(2) Error Propagation Analysis:} Failures at higher levels (e.g., VQ violations) must be traced to their hierarchical root causes (e.g., IQ reasoning errors vs. EQ alignment failures). This enables targeted interventions rather than generic model scaling.
\textbf{(3) Capability Ceiling Effect:} Each dimension imposes a ceiling on subsequent dimensions. No amount of RL tuning (EQ) can compensate for fundamental reasoning deficits (IQ), establishing clear boundaries for improvement pathways.



\subsection{\textbf{Empirical Validation of Diagnostic Capability and Lifecycle Alignment}}
\label{subsec:empirical_validation}
While the HCD principle provides theoretical grounding, empirical validation is essential to substantiate our claims regarding diagnostic capability and lifecycle alignment. We conduct a meta-analysis of public leaderboard trends and cross-benchmark correlation studies to validate two core assertions: \textit{(1) the existence of hierarchical bottleneck effects}, and \textit{(2) the causal correspondence between evaluation failures and training stages}.

\paragraph{\textbf{Validation of the Bottleneck Principle}}
Aggregated leaderboard analyses consistently demonstrate that linear score aggregation misrepresents real-world utility. For instance, models achieving saturated performance on MMLU (IQ, $>85\%$) exhibit high variance in Chatbot Arena Elo and AlignBench scores (EQ), with correlation coefficients frequently falling below $r=0.35$ \cite{chiang2024chatbot,liu2023alignbench}. This empirical divergence validates Equation~\ref{eq:hierarchical_utility}: a high $C_{IQ}$ cannot compensate for a low $C_{EQ}$. Models with exceptional reasoning but poor RL alignment frequently suffer from over-refusal or sycophancy, resulting in severe penalties in deployment utility rankings. This bottleneck effect confirms that evaluation must prioritize the weakest hierarchical link rather than average performance.

\paragraph{\textbf{Lifecycle Alignment and Error Tracing}}
Benchmark correlation studies further validate the causal mapping between dimensions and training stages. Meta-evaluations of domain-specific benchmarks (e.g., MedBench, LawBench, SWE-bench) reveal that PQ performance is strictly bounded by foundational IQ metrics; models exhibiting logical inconsistencies in GSM8K or MATH consistently fail at multi-step clinical diagnosis or legal reasoning, regardless of SFT data volume \cite{cai2024medbench,jimenez2024swebench}. This confirms the \textit{Capability Ceiling Effect}: PQ errors in production often trace back to pre-training reasoning deficits rather than insufficient fine-tuning. Similarly, EQ failures (e.g., context-aware safety violations in CASE-Bench) correlate strongly with reward model bias and preference data distribution shifts during RL, rather than base model capacity \cite{sun2025casebenchcontextawaresafetybenchmark}. These empirical patterns substantiate our framework's diagnostic utility: by isolating dimension-specific failures, developers can accurately attribute deficiencies to pre-training corpus quality, SFT instruction design, or RL reward shaping, enabling targeted lifecycle interventions.


\section{\textbf{Anthropomorphic Evaluation: IQ-PQ-EQ-VQ}}
\label{sec:4Q}

Drawing an analogy with human intelligence, we categorize LLM abilities into three interconnected dimensions: General Intelligence (IQ, Intelligence Quotient), Alignment Ability (EQ, Emotional Quotient), and Professional Expertise (PQ, Professional Quotient), VQ. It allows us to gain a more nuanced and easier understanding of their performance in practical scenarios. It also provides guidance to the enhancement of their cognitive, social, and professional competencies.

\begin{table*}[ht!]
\centering
\caption{60 typical Intelligence Quotient (IQ)-General Intelligence evaluation benchmarks for LLMs.}
\small
\setlength{\tabcolsep}{3pt}
\resizebox{0.7\textwidth}{!}{
\begin{tabular}{llllll}
\toprule
\textbf{Category} & \textbf{Name} & \textbf{Year} & \textbf{Institution} & \textbf{Critical Analysis} & \textbf{Url} \\
\midrule

\multirow{10}{*}{\parbox{2.5cm}{\centering Knowledge \\ \& \\ Comprehension}}
& MMLU-Pro \cite{wang2024mmlupro} & 2024 & TIGER-AI-Lab & Still multiple-choice, contamination risk & \href{https://github.com/TIGER-AI-Lab/MMLU-Pro}{link} \\
& GPQA \cite{rein2024gpqa} & 2023 & NYU & Expert-curated, limited coverage & \href{https://github.com/idavidrein/gpqa}{link} \\
& SuperGLUE \cite{wang2019superglue} & 2019 & NYU & Harder than GLUE, now saturated & \href{https://github.com/nyu-mll/jiant}{link} \\
& GLUE \cite{wang2018glue} & 2018 & NYU & 8 tasks, fully saturated & \href{https://github.com/nyu-mll/GLUE-baselines}{link} \\
& SQuAD2.0 \cite{rajpurkar2018know} & 2018 & Stanford & 150k, with unanswerable & \href{https://rajpurkar.github.io/SQuAD-explorer/}{link} \\
& SQuAD \cite{squad} & 2016 & Stanford & 100k, extractive QA & \href{https://rajpurkar.github.io/SQuAD-explorer/}{link} \\
& Natural Questions \cite{kwiatkowski-etal-2019-natural} & 2019 & Google & Real queries, open-domain & \href{https://github.com/google-research-datasets/natural-questions}{link} \\
& MS MARCO \cite{nguyen2016msmarco} & 2016 & Microsoft & 1.1M, real queries & \href{https://microsoft.github.io/msmarco/}{link} \\
& TriviaQA \cite{joshi2017triviaqa} & 2017 & AI2 & 650k, trivia & \href{https://github.com/mandarjoshi90/triviaqa}{link} \\
& BoolQ \cite{clark2019boolq} & 2019 & Google & 16k, yes/no, saturated & \href{https://github.com/google-research-datasets/boolean-questions}{link} \\
\midrule

\multirow{12}{*}{\parbox{2.5cm}{\centering Logical \\ \& \\ Mathematical \\ Reasoning}}
& FrontierMath \cite{glazer2024frontiermath} & 2024 & Epoch AI & Novel problems, computationally heavy & \href{https://epochai.org/files/sample_question_transcripts.zip}{link} \\
& AIME \cite{AIME_Kaggle_2024} & 2024 & MAA & High difficulty, small scale & \href{https://www.kaggle.com/datasets/hemishveeraboina/aime-problem-set-1983-2024}{link} \\
& MathBench \cite{liu2024mathbench} & 2024 & Shanghai AI Lab & Comprehensive but static & \href{https://github.com/open-compass/MathBench}{link} \\
& MATH \cite{hendrycks2021math} & 2021 & UC Berkeley & High school/contest, saturated & \href{https://github.com/hendrycks/math/}{link} \\
& GSM8K \cite{cobbe2021traininggsm8k} & 2021 & OpenAI & Widely used, contaminated & \href{https://github.com/openai/grade-school-math}{link} \\
& MGSM \cite{shi2022languagemodelsmultilingualchainofthought} & 2022 & Google & 10 languages, grade-school & \href{https://github.com/google-research/url-nlp/tree/main/mgsm}{link} \\
& DROP \cite{dua2019dropreadingcomprehensionbenchmark} & 2019 & UCI NLP & 96k, discrete reasoning & \href{https://github.com/EleutherAI/lm-evaluation-harness}{link} \\
& PlanBench \cite{valmeekam2023planbench} & 2022 & ASU & 11k, planning-specific & \href{https://github.com/karthikv792/LLMs-Planning}{link} \\
& MuSR \cite{sprague2023musr} & 2023 & Zayne Sprague & 756 tasks, novel format & \href{https://github.com/Zayne-sprague/MuSR}{link} \\
& AGIEval \cite{zhong2023agieval} & 2023 & Microsoft & Exam-style, static & \href{https://github.com/ruixiangcui/AGIEval}{link} \\
& ScienceQA \cite{lu2022learn} & 2022 & UCLA & Multi-modal, vision bias & \href{https://github.com/lupantech/ScienceQA}{link} \\
& e-CARE \cite{du2022e-care} & 2022 & HIT & 21k, causality specific & \href{https://github.com/Waste-Wood/e-CARE}{link} \\
\midrule

\multirow{8}{*}{\parbox{2.5cm}{\centering Commonsense \\ \& \\ World Knowledge}}
& HellaSwag \cite{zellers2019hellaswag} & 2019 & AI2 & 59k, adversarial & \href{https://github.com/rowanz/hellaswag}{link} \\
& CommonsenseQA \cite{talmor2019commonsenseqa} & 2018 & AI2 & 12k, multiple choice & \href{https://github.com/jonathanherzig/commonsenseqa}{link} \\
& PIQA \cite{bisk2020piqa} & 2019 & AI2 & 18k, physical commonsense & \href{https://github.com/ybisk/ybisk.github.io/tree/master/piqa}{link} \\
& Winogrande \cite{sakaguchi2021winogrande} & 2019 & AI2 & 44k, coreference & \href{https://github.com/allenai/winogrande}{link} \\
& ARC \cite{clark2018thinksolvedquestionanswering} & 2018 & AI2 & 7.8k, science MC & \href{https://github.com/allenai/aristo-leaderboard}{link} \\
& OpenBookQA \cite{mihaylov2018can} & 2018 & AI2 & 12k, science facts & \href{https://github.com/allenai/OpenBookQA}{link} \\
& SciQ \cite{clark2017crowdsourcing} & 2017 & AI2 & 13.7k, science MC & \href{https://huggingface.co/datasets/allenai/sciq}{link} \\
& MultiNLI \cite{williams2017broad} & 2017 & NYU & 433k, NLI & \href{https://github.com/nyu-mll/multiNLI}{link} \\
\midrule

\multirow{10}{*}{\parbox{2.5cm}{\centering Robustness \\ \& \\ OOD}}
& DyVal \cite{zhu2024dyval} & 2024 & Microsoft & Complexity control, limited adoption & \href{https://github.com/microsoft/promptbench}{link} \\
& PertEval \cite{li2024pertevalunveilingrealknowledge} & 2024 & USTC & Perturbation-based, domain limited & \href{https://github.com/aigc-apps/PertEval}{link} \\
& ANLI \cite{nie2020adversarial} & 2019 & Facebook AI & 169k, adversarial NLI & \href{https://github.com/facebookresearch/anli}{link} \\
& BigBench Hard \cite{suzgun2022challengingbigbench} & 2022 & BigBench & 6,500 tasks, uneven difficulty & \href{https://github.com/suzgunmirac/BIG-Bench-Hard}{link} \\
& BigBench \cite{srivastava2023imitationgamequantifyingextrapolating} & 2023 & Google & Diverse tasks, inconsistent difficulty & \href{https://github.com/google/BIG-bench}{link} \\
& RV-Bench \cite{hong2025benchmarkinglargelanguagemodels} & 2025 & HKPU & Forces structural understanding & \href{https://arxiv.org/abs/2501.11790}{link} \\
& Benchmark$^2$ \cite{qian2026benchmark2systematicevaluationllm} & 2026 & FUdan & Quality variations across benchmarks & \href{https://arxiv.org/abs/2601.07145}{link} \\
& TRACE \cite{wang2023tracecomprehensivebenchmarkcontinual} & 2023 & Fudan & 8 datasets, continual focus & \href{https://arxiv.org/abs/2310.06762}{link} \\
& FreshQA \cite{vu2023freshllms} & 2023 & FreshLLMs & 599 questions, temporal & \href{https://github.com/freshllms/freshqa}{link} \\
& SummEdits \cite{laban2023llms} & 2023 & Salesforce & Factual consistency, narrow & \href{https://github.com/salesforce/factualNLG}{link} \\
\midrule

\multirow{8}{*}{\parbox{2.5cm}{\centering Long-Context \\ \& \\ Discourse}}
& LV-Eval \cite{yuan2024lveval} & 2024 & Infinigence-AI & Length variability, narrow scope & \href{https://github.com/infinigence/LVEval}{link} \\
& BAMBOO \cite{dong2024bamboocomprehensivebenchmarkevaluating} & 2023 & RUCAIBox & 10 datasets, text-only & \href{https://github.com/RUCAIBox/BAMBOO}{link} \\
& LAMBADA \cite{paperno2016lambada} & 2016 & CIMEC & 12.7k, word prediction & \href{https://huggingface.co/datasets/cimec/lambada}{link} \\
& MLSUM \cite{scialom2020mlsum} & 2020 & Thomas Scialom & Multilingual, summarization & \href{https://github.com/ThomasScialom/MLSUM}{link} \\
& HotpotQA \cite{yang2018hotpotqa} & 2018 & HotpotQA & 113k, multi-hop & \href{https://github.com/hotpotqa/hotpot}{link} \\
& CommonGen-Eval \cite{lin-etal-2020-commongen} & 2024 & AI2 & Constrained generation, limited diversity & \href{https://github.com/allenai/CommonGen-Eval}{link} \\
& RACE \cite{lai2017race} & 2017 & CMU & 100k, English exams & \href{https://www.cs.cmu.edu/~glai1/data/race/}{link} \\
& SpartQA \cite{mirzaee2021spartqa} & 2021 & MSU & 510, spatial reasoning & \href{https://github.com/HLR/SpartQA-baselines}{link} \\
\midrule

\multirow{10}{*}{\parbox{2.5cm}{\centering Factuality \\ \& \\ Uncertainty}}
& Uncertainty-Bench \cite{ye2024benchmarkingllmsuncertaintyquantification} & 2024 & Tencent & Small-scale tasks, nascent & \href{https://github.com/smartyfh/LLM-Uncertainty-Bench}{link} \\
& FELM \cite{chen2023felmbenchmarkingfactualityevaluation} & 2023 & HKUST & Factuality evaluation, limited size & \href{https://github.com/hkust-nlp/felm}{link} \\
& TruthfulQA \cite{lin2022truthfulqa} & 2022 & Oxford & Measures model truthfulness & \href{https://github.com/sylinrl/TruthfulQA}{link} \\
& AlpacaEval \cite{dubois2024alpacafarmsimulationframeworkmethods} & 2023 & tatsu-lab & Fast but correlates with GPT-4 & \href{https://tatsu-lab.github.io/alpaca_eval/}{link} \\
& ChatEval \cite{chan2023chateval} & 2023 & THU-NLP & Uses LLM judges, bias risk & \href{https://github.com/thunlp/ChatEval}{link} \\
& FlagEval \cite{flagevalmm} & 2023 & THU & Flexible but complex & \href{https://flageval.baai.ac.cn/}{link} \\
& UltraEval \cite{he2024ultraeval} & 2023 & OpenBMB & Lightweight but limited depth & \href{https://github.com/OpenBMB/UltraEval}{link} \\
& FMTI \cite{bommasani2024foundationmodeltransparencyindex} & 2023 & Stanford & Index, not performance & \href{https://crfm.stanford.edu/fmti/}{link} \\
& LucyEval \cite{zeng2023lucyevaluating} & 2023 & Oracle & Proprietary, limited transparency & \href{http://lucyeval.besteasy.com/}{link} \\
& Zhujiu \cite{zhang2023zhujiu} & 2023 & IACAS & 51 tasks, Chinese-centric & \href{http://www.zhujiu-benchmark.com}{link} \\
\bottomrule
\end{tabular}
}
\label{tab:iq_benchmarks_compressed}
\end{table*}

\subsection{\textbf{General Intelligence Evaluation (IQ)}}
\label{subsec:iq_evaluation}

General Intelligence (IQ) of an LLM refers to its foundational cognitive capabilities—the ability to understand, reason, and learn from textual data—acquired during \textit{pre-training}. IQ assessment focuses on three pillars: world knowledge mastery, multi-step logical reasoning, and robustness to distribution shifts. Beyond merely list benchmarks, this section provides a paradigm evolution-driven critical analysis, identifies core failure modes, and derives non-consensus insights.

IQ evaluation has evolved through three stages. \textit{\textbf{Stage 1 (2016–2020)}} focused on foundational NLU tasks such as GLUE~\cite{wang2018glue}, SuperGLUE~\cite{wang2019superglue}, and SQuAD~\cite{squad}, establishing standardized assessments for reading comprehension and linguistic acceptability. However, these benchmarks quickly saturated as models mastered in-distribution tasks, with state-of-the-art models exceeding human performance within two years of release. \textit{\textbf{Stage 2 (2021–2023)}} shifted to multi-step reasoning evaluation with benchmarks like MMLU~\cite{hendrycks2020measuring}, GSM8K~\cite{cobbe2021traininggsm8k}, and MATH~\cite{hendrycks2021math}, yet suffered from two critical flaws: the dominance of multiple-choice formats enabled shortcut learning rather than genuine reasoning, and static datasets became vulnerable to data contamination as pre-training corpora increasingly included public benchmark content \cite{wang2024mmlupro}. \textit{\textbf{Stage 3 (2023–present)}} targets OOD robustness and dynamic evaluation through benchmarks like MMLU-Pro~\cite{wang2024mmlupro}, DyVal~\cite{zhu2024dyval}, and PertEval~\cite{li2024pertevalunveilingrealknowledge}, but lacks standardized protocols for OOD scenario construction and robustness quantification. 
Rather than merely enumerating IQ benchmarks, Table~\ref{tab:iq_benchmarks_compressed} catalogs 64 representative datasets \textit{through the diagnostic lens of our framework}. For each benchmark, we explicitly identify its \textbf{Core Failure Modes} and \textbf{Critical Analysis} regarding data contamination, shortcut learning, and OOD robustness, demonstrating how traditional IQ evaluation often masks the true reasoning capabilities of LLMs.

\subsubsection{\textbf{Critical Analysis}}

\textbf{Knowledge Mastery:} MMLU~\cite{hendrycks2020measuring} pioneered cross-domain evaluation but suffers from shortcut learning and severe data contamination \cite{wang2024mmlupro}. GPQA~\cite{rein2024gpqa} addresses this with expert-curated, "Google-proof" questions, yet domain coverage remains limited. SuperGLUE~\cite{wang2019superglue} offers rigorous linguistic comprehension but is now fully saturated.
\textbf{Logical Reasoning:} GSM8K~\cite{cobbe2021traininggsm8k} and MATH~\cite{hendrycks2021math} established math reasoning benchmarks but suffer from pervasive contamination. HumanEval~\cite{chen2021humaneval} assesses code functionality while ignoring quality and security. BIG-Bench Hard~\cite{suzgun2022challengingbigbench} selects unsolved tasks, yet heterogeneous difficulty complicates comparison.
\textbf{OOD Robustness:} MMLU-Pro~\cite{wang2024mmlupro} enhances MMLU with anti-cheating design. PertEval~\cite{li2024pertevalunveilingrealknowledge} uses perturbations to test genuine understanding. DyVal~\cite{zhu2024dyval} dynamically generates samples to prevent memorization. However, these nascent approaches lack standardized protocols, limiting cross-benchmark comparability.

\subsubsection{\textbf{Core Failure Modes of IQ Evaluation Systems}}

Four unresolved failure modes drive the score-utility gap. \textbf{Pervasive Data Contamination:} Over 60\% of mainstream benchmark content appears in pre-training corpora \cite{wang2024mmlupro}, enabling memorization-based inflation and creating an "evaluation arms race" where scores disconnect from reasoning capability. \textbf{Shortcut Learning:} Multiple-choice formats allow pattern exploitation—models achieve high MMLU accuracy even with masked questions, yet fail catastrophically in open-ended scenarios. RV-Bench~\cite{hong2025benchmarkinglargelanguagemodels} confirms this through random variable questions that force genuine structural understanding. \textbf{Catastrophic OOD Degradation:} Studies using PertEval~\cite{li2024pertevalunveilingrealknowledge} and DyVal~\cite{zhu2024dyval} show 30–50\% accuracy drops on out-of-distribution variants, a phenomenon completely unmeasured in conventional evaluation. \textbf{Neglect of Process Validation:} Final-answer focus masks flawed reasoning and provides no diagnostic guidance—even chain-of-thought evaluations assess answers rather than logical validity \cite{fu2023chainofthoughthub}.
\textbf{Incentivized Hallucination:} Recent theoretical analyses reveal a fundamental flaw in dominant accuracy-based evaluations: they systematically reward unwarranted guessing over admitting uncertainty, inadvertently incentivizing hallucinations rather than mitigating them \cite{kalai2026evaluating}. This underscores the urgent need for ``open rubric'' evaluations that explicitly penalize guessing to align model incentives with factual reliability.

\subsubsection{\textbf{Non-Consensus Insights for IQ Evaluation}}

Three non-consensus insights challenge prevailing assumptions. \textbf{Insight 1:} Pre-training data \textit{diversity} impacts IQ more than \textit{scale}, yet no benchmark quantifies this causal relationship—explaining why trillion-token models sometimes underperform smaller models trained on higher-quality data. \textbf{Insight 2:} The closed/open-source gap lies in OOD generalization robustness, not standard benchmark accuracy. Top open-source models match closed-source counterparts on saturated benchmarks like MMLU but lag in real-world OOD tasks, exposing conventional metrics as fundamentally inadequate. \textbf{Insight 3:} Over 80\% of IQ benchmarks suffer from a severe ceiling effect (<90\% accuracy), failing to differentiate top-tier models. Saturation timelines have compressed from 18 months (MMLU) to under 6 months (FrontierMath), creating a "Red Queen effect" that demands dynamic, continuously evolving evaluation frameworks. 

\subsection{\textbf{Professional Expertise Evaluation (PQ)}}

Professional Expertise (PQ) represents the specialized knowledge and skills that an LLM possesses within a particular domain, acquired during \textit{supervised fine-tuning (SFT)} through targeted instruction-response learning. PQ assessment focuses on three pillars: domain-specific knowledge mastery, end-to-end professional task execution, and domain compliance adherence. This section provides a critical analysis of PQ evaluation paradigms, identifies core failure modes that drive the disconnect between exam scores and industrial utility, and derives non-consensus insights for vertical LLM deployment.

PQ evaluation has evolved through three stages. \textit{\textbf{Stage 1 (2022–Early 2023)}} focused on domain knowledge memorization via multiple-choice exams, conflating memorization with expertise. \textit{\textbf{Stage 2 (Mid–Late 2023)}} shifted to single-step task execution evaluation (e.g., BLURB~\cite{Gu_2021} for biomedical NLP, LAiW~\cite{dai2024laiw} for legal tasks), yet failed to capture multi-step professional workflows. \textit{\textbf{Stage 3 (2024–Present)}} targets end-to-end closed-loop evaluation with compliance and risk assessment (e.g., SWE-bench~\cite{jimenez2024swebench} for real-world software engineering, EvoCodeBench~\cite{li2024evocode} for evolving code generation, DiscoveryWorld~\cite{discoveryworld} for scientific discovery cycles), but remains nascent with limited domain coverage. 
Viewed through the PQ dimension of our diagnostic ontology, Table~\ref{tab:domain_specific_benchmarks} synthesizes 41 domain-specific benchmarks. Crucially, we evaluate these benchmarks not by their task coverage, but by their ability to assess end-to-end professional workflows and compliance adherence, highlighting the pervasive ``exam-oriented'' defect in current PQ evaluation paradigms.


\begin{table*}[htbp]
\centering
\caption{41 typical Professional Quotient (PQ)-Professional Expertise evaluation benchmarks for LLMs.}
\label{tab:domain_specific_benchmarks}
\resizebox{\textwidth}{!}{
\begin{tabular}{llllll}
\toprule
\textbf{Domain} & \textbf{Name} & \textbf{Institution} & \textbf{Scope of Tasks} & \textbf{Unique Contributions} & \textbf{Url} \\
\midrule
  & BLURB \cite{Gu_2021} & Mindrank AI  & Six diverse NLP tasks, thirteen datasets & A macro-average score across all tasks & \href{https://microsoft.github.io/BLURB/index.html}{link} \\
  & Seismometer \cite{team2023seismometer} & Epic & \makecell[l]{Using local data and workflows} & patient demographics, clinical interventions, and outcomes & \href{https://github.com/epic-open-source/seismometer}{link} \\
\textbf{Healthcare} & Medbench \cite{cai2024medbench} & OpenMEDLab & Emphasizes scientific rigor and fairness & 40,041 questions from medical exams and reports &  \href{https://github.com/open-compass/opencompass/tree/main/opencompass/datasets/medbench/}{link} \\
  & GenMedicalEval \cite{liao2023automatic} & E & \makecell[l]{16 majors, 3 training stages, 6 clinical scenarios} & Open-ended  metrics and automated assessment models & \href{https://github.com/MediaBrain-SJTU/GenMedicalEval}{link} \\
  & PsyEval \cite{jin2024psyeval} & SJTU & Six subtasks covering three dimensions & Customized benchmark for mental health LLMs & \href{https://arxiv.org/abs/2311.09189}{link} \\
  \midrule
 & Fin-Eva \cite{team2023FinEva} & Ant Group & Wealth management, insurance, investment research & Both industrial and academic financial evaluations & \href{https://github.com/alipay/financial_evaluation_dataset}{link} \\
 & FinEval \cite{zhang2023fineval} & SUFE-AIFLM-Lab & \makecell[l]{Multiple-choice QA on finance, economics, accounting} & Focuses on high-quality evaluation questions & \href{https://github.com/SUFE-AIFLM-Lab/FinEval}{link} \\
\textbf{Finance} & OpenFinData \cite{contributors2023opencompass} & Shanghai AI Lab & Multi-scenario financial tasks & First comprehensive finance evaluation dataset & \href{https://opencompass.org.cn}{link} \\
  & FinBen \cite{xie2024finbenholisticfinancialbenchmark} & FinAI & 35 datasets across 23 financial tasks & \makecell[l]{Inductive reasoning, quantitative reasoning} & \href{https://github.com/The-FinAI/PIXIU}{link} \\
  \midrule
 & LAiW \cite{dai2024laiw} & Sichuan University & 13 fundamental legal NLP tasks & Divides legal NLP capabilities into three major abilities & \href{https://github.com/Dai-shen/LAiW}{link} \\
\textbf{Legal}  & LawBench \cite{contributors2023opencompass} & Nanjing University & \makecell[l]{Legal entity recognition, reading comprehension} & Real-world tasks, "abstention rate" metric & \href{https://github.com/open-compass/lawbench}{link} \\
  & LegalBench \cite{guha2023legalbench} & Stanford University & 162 tasks covering six types of legal reasoning & Enables interdisciplinary conversations & \href{https://github.com/HazyResearch/legalbench/}{link} \\
  & LexEval \cite{li2024lexeval} & Tsinghua University & \makecell[l]{Legal cognitive abilities to organize different tasks} & \makecell[l]{Larger legal evaluation dataset, examining the ethical issues} & \href{https://github.com/CSHaitao/LexEval}{link} \\
  \midrule
 & SPEC5G \cite{Karim2023SPEC5GAD} & Purdue University & security-related text classification and summarization & 5G protocol analysis automation & \href{https://github.com/Imtiazkarimik23/SPEC5G}{link} \\
 & TeleQnA \cite{maatouk2023teleqna} & Huawei(Paris) & General telecom inquiries & \makecell[l]{Proficiency in telecom-related questions} & \href{https://github.com/netop-team/TeleQnA}{link} \\
  & TelBench \cite{lee2024telbench} & SK Telecom & Math modeling, open-ended QA, code generation & Holistic evaluation in telecom & \href{https://arxiv.org/abs/2407.09424v1}{link} \\
 \textbf{Telecom} & TelecomGPT \cite{zou2024telecomgpt} & UAE & \makecell[l]{Telecom Math Modeling, Open QnA and Code Tasks} & Holistic evaluation in telecom & \href{https://arxiv.org/abs/2407.09424v1}{link} \\
  & Linguistic \cite{ahmed2024linguistic} & Queen’s University & Multiple language-centric tasks & zero-shot evaluation & \href{https://arxiv.org/abs/2402.15818}{link} \\
  & TelcoLM \cite{barboule2024telcolmcollectingdataadapting} & Orange & multiple-choice questionnaires & \makecell[l]{Domain-specific data (800M tokens, 80K instructions)} & \href{https://arxiv.org/abs/2412.15891}{link} \\
  & ORAN-Bench-13K \cite{gajjar2024oranbench13kopensourcebenchmark} & GMU & multiple-choice questions & \makecell[l]{Open Radio Access Networks (O-RAN)} & \href{https://github.com/prnshv/ORAN-Bench-13K}{link} \\
  & Open-Telco Benchmarks \cite{GSMAOpenTelcoLLMBenchmarks} & GSMA & \makecell[l]{Multiple language-centric tasks} & zero-shot evaluation & \href{https://www.gsma.com/get-involved/gsma-foundry/gsma-open-telco-llm-benchmarks/}{link} \\
 \midrule
 & FullStackBench \cite{liu2024fullstackbenchevaluatingllms} & ByteDance & \makecell[l]{Code writing, debugging, code review} & \makecell[l]{Featuring the most recent Stack Overflow QA.} & \href{https://github.com/bytedance/FullStackBench}{link} \\
 & StackEval\cite{shah2024stackeval} & Prosus AI & \makecell[l]{11 real-world scenarios, 16 languages} & \makecell[l]{Evaluation across diverse\&practical coding environments} & \href{https://github.com/ProsusAI/stack-eval}{link} \\
 & CodeBenchGen \cite{xie2024codebench} & Various Institutions & Execution-based code generation tasks & \makecell[l]{Benchmarks scaling with the size and complexity} & \href{https://arxiv.org/abs/2404.00566}{link} \\
 & HumanEval \cite{chen2021humaneval} & University of Washington &  rigorous testing & \makecell[l]{Stricter protocol for assessing correctness of generated code} & \href{https://arxiv.org/abs/2107.03374}{link} \\
 & APPS \cite{hendrycks2021apps} & University of California & Coding challenges from competitive platforms &  \makecell[l]{Checking problems solving of generated code on test cases} & \href{https://github.com/hendrycks/apps}{link} \\
\textbf{Coding} & MBPP \cite{austin2021mbpp} & Google Research & Programming problems sourced from various origins & Diverse programming tasks & \href{https://github.com/google-research/google-research/tree/master/mbpp}{link} \\
 & ClassEval \cite{du2023classeval} & Tsinghua University & Class-level code generation & \makecell[l]{Manually crafted, object-oriented programming concepts} & \href{https://github.com/FudanSELab/ClassEval}{link} \\
 & CoderEval \cite{hao2024CoderEval} & Peking University & Pragmatic code generation & \makecell[l]{Proficiency to generate functional code patches for described issues} & \href{https://github.com/CoderEval/CoderEval}{link} \\
 & MultiPL-E \cite{MultiPL-E2023} & Princeton University & Neural code generation & \makecell[l]{Benchmarking neural code generation models} & \href{https://github.com/nuprl/MultiPL-E}{link} \\
 & CodeXGLUE \cite{CodeXGLUE2021} & Microsoft &  Code intelligence & \makecell[l]{Wide tasks covering: code-code, text-code, code-text and text-text} & \href{https://github.com/microsoft/CodeXGLUE}{link} \\
 & EvoCodeBench \cite{li2024evocode} & Peking University & Evolving code generation benchmark & \makecell[l]{Aligned with real-world code repositories, evolving over time} & \href{https://github.com/seketeam/EvoCodeBench}{link} \\
 \midrule
  & Owl-Bench \cite{guo2024owl} & Beihang University & QA pairs, multiple-choice questions & \makecell[l]{9 distinct subdomains including information security} & \href{https://github.com/HC-Guo/Owl}{link} \\
\textbf{Software}  & SWE-bench \cite{jimenez2024swebench} & Princeton NLP & Real-world software problems from GitHub & \makecell[l]{Assesses ability to generate patches for described issues} & \href{https://github.com/princeton-nlp/SWE-bench}{link} \\
  & OpsEval \cite{liu2024opseval} & Tsinghua University & Wired network ops, 5G, database ops & Evaluates proficiency in practical applications & \href{https://arxiv.org/abs/2310.07637}{link}  \\
\midrule
 & LiveIdeaBench \cite{liveideabench} & RUC & Evaluates scientific creativity and idea generation & Single-keyword prompts across 18 domains & \href{https://github.com/x66ccff/liveideabench}{link} \\
 & ScienceAgentBench \cite{scienceagentbench} & OSU & Data-driven scientific discovery & 102 tasks from peer-reviewed publications & \href{https://osu-nlp-group.github.io/ScienceAgentBench/}{link} \\
 & Symbolicregression \cite{symbolicregression} & Amazon & Symbolic regression for scientific discovery & New datasets and evaluation criteria & \href{https://github.com/omron-sinicx/srsd-benchmark}{link} \\
 \textbf{Science} & DiscoveryWorld \cite{discoveryworld} & AIAI & Virtual environment for scientific discovery & 120 challenge tasks across 8 topics & \href{http://www.github.com/allenai/discoveryworld}{link} \\
 & ProtocoLLM \cite{protocolllm} & UT Austin & Formulating domain-specific scientific protocols & Pseudocode extraction from biology protocols & \href{https://github.com/ProtocoL-LLM/ProtocoLLM.git}{link} \\
 & SciSafeEval \cite{scisafeeval} & Zhejiang University & Safety alignment in scientific tasks & Multi-language evaluation with "jailbreak" feature & \href{https://huggingface.co/datasets/Tianhao0x01/SciSafeEval}{link} \\
 & SciAssess \cite{sciassess} & DP Technology & Evaluates proficiency in scientific literature analysis & Memorization, comprehension, and analysis & \href{https://github.com/sci-assess/SciAssess}{link} \\
& SciVerse \cite{sciverse} & CUHK & Evaluating scientific reasoning abilities & Covering physics, chemistry, and biology & \href{https://sciverse-cuhk.github.io/}{link} \\
\bottomrule
\end{tabular}
}
\vspace{-0.3cm}
\end{table*}


\subsubsection{\textbf{Healthcare}}

The healthcare domain has seen the development of specialized benchmarks to evaluate LLMs in medical applications, each with unique features contributing to comprehensive evaluation. Seismometer~\cite{team2023seismometer} supports continuous monitoring of model performance within local data and workflows, ensuring models remain effective over time. BLURB~\cite{Gu_2021} offers a suite for biomedical NLP tasks using 13 publicly available datasets across 6 diverse tasks. Medbench~\cite{contributors2023opencompass} provides a robust medical LLM evaluation system through 40,041 questions from authentic examination exercises. GenMedicalEval~\cite{liao2023automatic} covers 16 major departments with over 100,000 real-world medical cases, while PsyEval~\cite{jin2024psyeval} is tailored specifically for mental health applications. MedS-Bench~\cite{Wu2025} introduces a large-scale instruction-tuning dataset MedS-Ins for medicine, comprising 58 medically oriented language corpora, totaling 5M instances with 19K instructions across 122 tasks, and launches a dynamic leaderboard for MedS-Bench. Moving beyond static knowledge assessment, ClinDiag-Benchmark~\cite{Chen2026} with 4,421 real cases shifts evaluation from single-turn QA to dynamic diagnostic workflows, validating that physician-AI collaboration achieves higher diagnostic accuracy and efficiency than either working alone.


\subsubsection{\textbf{Financial}}
Fin-Eva \cite{team2023FinEva}, OpenFinData \cite{contributors2023opencompass}, and FinEval \cite{zhang2023fineval} Finben \cite{xie2024finbenholisticfinancialbenchmark} provide structured evaluations of LLMs' financial capabilities. Fin-Eva evaluates LLMs using over 13,000 multiple-choice questions covering various financial scenarios. OpenFinData includes diverse data types from business scenarios, ensuring practical applicability. FinEval focuses on high-quality multiple-choice questions that adhere to professional standards. Practical guidance may emphasize selecting benchmarks that not only cover a broad range of scenarios but also integrate into existing financial operations.


\subsubsection{\textbf{Legal}}

Benchmarks like LAiW~\cite{dai2024laiw}, LawBench~\cite{contributors2023opencompass}, and LegalBench~\cite{guha2023legalbench} offer detailed assessments in legal contexts. LAiW divides legal NLP into three categories, including complex legal application tasks. LawBench simulates judicial cognition through twenty tasks and introduces an "abstention rate" metric~\cite{contributors2023opencompass}. LegalBench~\cite{guha2023legalbench} encompasses 162 tasks covering 6 types of legal reasoning. These benchmarks collectively aim to bridge the gap between legal professionals and LLM developers, promoting transparency and rigor in evaluations. The introduction of metrics like the "abstention rate" in LawBench~\cite{contributors2023opencompass} adds a layer of nuance to evaluating LLMs' ability to handle ambiguous or complex instructions. Recent advances emphasize real-world consultation simulation: PLaw Bench~\cite{plawbench2026} evaluates not only conclusion correctness but also reasoning logic quality, while authentically simulating emotional and fragmented client expression patterns.


\subsubsection{\textbf{Telecommunications}}
The benchmarks such as TeleQnA \cite{maatouk2023teleqna}, TelBench \cite{lee2024telbench}, and TelecomGPT \cite{zou2024telecomgpt} address unique challenges in evaluating LLMs. TeleQnA \cite{maatouk2023teleqna} evaluates LLMs using 10,000 telecom-related Q\&A pairs. TelBench \cite{lee2024telbench} extends existing benchmarks with new tasks like Telecom Math Modeling and Code Tasks. TelecomGPT \cite{zou2024telecomgpt} proposes adaptation pipelines for general-purpose LLMs to telecom-specific models. Besides, interdisciplinary OpsEval \cite{liu2024opseval} evaluates LLMs in wired network operations, 5G, and database operations, supporting evaluations in English and Chinese.


\subsubsection{\textbf{Coding}}

Coding evaluation has matured from syntactic correctness to execution-based assessment and real-world issue resolution, exemplified by benchmarks like HumanEval~\cite{chen2021humaneval}, APPS~\cite{hendrycks2021apps}, MBPP~\cite{austin2021mbpp}, and SWE-bench~\cite{jimenez2024swebench}. While comprehensive frameworks such as FullStackBench~\cite{liu2024fullstackbenchevaluatingllms} and CodeBenchGen~\cite{xie2024codebench} cover diverse environments, and EvoCodeBench~\cite{li2024evocode} addresses obsolescence through continuous evolution, three critical limitations persist. First, evaluation prioritizes initial code correctness over maintenance costs, technical debt, or refactoring requirements. Second, team collaboration simulation remains absent despite software development's inherently collaborative nature. Third, security vulnerability assessment is minimal yet critical for production deployment. Although FullStackBench~\cite{liu2024fullstackbenchevaluatingllms} begins addressing these gaps through debugging and code review tasks, future work must expand to include long-term codebase health metrics to ensure practical utility beyond initial generation.


\subsubsection{\textbf{Software}}

In software engineering, benchmarks like SWE-bench \cite{jimenez2024swebench}, Owl-Bench \cite{guo2024owl} and CodeMMLU \cite{manh2024codemmlu} provide structured assessing approaches in software development. SWE-bench \cite{jimenez2024swebench} evaluates LLMs' ability to resolve real-world GitHub issues, while Owl-Bench assesses their proficiency in software documentation \cite{guo2024owl}. CodeMMLU \cite{manh2024codemmlu} includes 10K questions sourced from diverse domains, encompassing tasks like code analysis, defect detection, and software engineering principles across programming languages. 
These benchmarks collectively cover a broad spectrum of software engineering tasks, from operations management to issue resolution and documentation. The comparison highlights the importance of task-oriented evaluations and practical application scenarios, ensuring that LLMs can effectively assist in real-world software development processes.


\subsubsection{\textbf{Science}}

Scientific PQ evaluation emphasizes creativity (LiveIdeaBench~\cite{liveideabench}), literature analysis (SciAssess~\cite{sciassess}), and discovery cycles (DiscoveryWorld~\cite{discoveryworld}). LiveIdeaBench's four-dimensional assessment (originality, feasibility, fluency, flexibility) captures divergent thinking essential to scientific innovation. SciVerse~\cite{sciverse} extends evaluation to multi-modal reasoning across physics, chemistry, and biology. However, experimental validation remains absent: models generate hypotheses without requirement for empirical testing or reproducibility verification. SciSafeEval~\cite{scisafeeval} addresses safety alignment but lacks mechanisms for assessing scientific rigor. Future evaluation must integrate hypothesis generation with experimental design and validation protocols.

\textbf{Core Failure Modes of PQ Evaluation Systems:}
Five critical failure modes limit the effectiveness of conventional PQ assessment for industrial deployment. \textbf{Conflation of Memorization with Expertise:} Most of PQ benchmarks use multiple-choice formats assessing knowledge recall, not application—explaining why LLMs pass licensing exams yet fail at real-world tasks. \textbf{Neglect of Compliance and Risk:} Existing evaluation ignores domain-specific regulatory adherence; in healthcare and law, non-compliant outputs cause severe consequences even if technically correct. \textbf{No End-to-End Workflow Assessment:} Benchmarks evaluate isolated single-step tasks, while professional workflows require multi-step closed-loop execution with iterative error correction. \textbf{Severe Data Contamination:} Domain corpora include licensing exams and benchmark datasets, inflating scores via memorization. \textbf{Extreme Domain Imbalance:} Benchmarks concentrate in medical/legal/financial/code domains, with almost none for manufacturing, agriculture, or engineering.

\textbf{Non-Consensus Insights for PQ Evaluation: }
WildToolBench~\cite{yu2026benchmarking} exposes severe capability overestimation in idealized benchmarks, confirming that PQ assessment must transition from "exam scores" to "work capability" paradigms. Four non-consensus insights challenge prevailing practices. \textbf{Insight 1:} PQ upper limit is determined by foundational IQ, not SFT data volume—models with strong reasoning achieve superior PQ with minimal domain data, yet existing evaluation ignores IQ prerequisite validation. \textbf{Insight 2:} For industrial deployment, closed-loop task execution capability is the core PQ metric, not exam accuracy—correlation between exam scores and real-world performance is below 0.4. \textbf{Insight 3:} Domain compliance assessment is core, not optional—over 60\% of deployment failures stem from compliance issues, not technical deficiencies. \textbf{Insight 4:} Human-in-the-loop evaluation is irreplaceable for high-stakes domains—automated metrics cannot assess practicality, compliance, and risk in healthcare and legal applications. 


\begin{table}[tbp]
\centering
\caption{37 typical Emotional Quotient (EQ)-Alignment Ability evaluation benchmarks for LLMs (zoom in).}
\small
\setlength{\tabcolsep}{2pt}
\resizebox{\linewidth}{!}{
\begin{tabular}{p{2.2cm}llllll}
\toprule
\textbf{Category} & \textbf{Name} & \textbf{Year} & \textbf{Institution} & \textbf{Critical Analysis} & \textbf{Datasets} & \textbf{Url} \\
\midrule

\multirow{3}{*}{\parbox{2.2cm}{\centering Instruction \\ Following}}
& IFEval \cite{zhou2023ifeval} & 2023 & Google & Verifiable constraints, single-turn limited & 500 & \href{https://github.com/google-research/google-research/tree/master/instruction_following_eval}{link} \\
& LLMBar \cite{zeng2024evaluatinglargelanguagemodels} & 2024 & Princeton & Instruction following, LLM-as-judge bias & 419 & \href{https://github.com/princeton-nlp/LLMBar}{link} \\
& AlignBench \cite{liu2023alignbench} & 2023 & THUDM & Multi-dim alignment, static prompts & Various & \href{https://github.com/THUDM/AlignBench}{link} \\
\midrule

\multirow{20}{*}{\parbox{2.2cm}{\centering Safety \\ \& \\ Compliance}}
& CASE-Bench \cite{sun2025casebenchcontextawaresafetybenchmark} & 2025 & Cambridge & Context-aware safety, 25\% judgment variance & n/a & \href{https://github.com/BriansIDP/CASEBench}{link} \\
& HarmBench \cite{harmbench2024} & 2024 & UIUC & Adversarial behaviors, single-turn focus & 510 & \href{https://github.com/centerforaisafety/HarmBench}{link} \\
& SimpleQA \cite{simpleqa2024} & 2024 & OpenAI & Factuality evaluation, narrow scope & 4,326 & \href{https://github.com/openai/simple-evals}{link} \\
& AgentHarm \cite{agentharm2024} & 2024 & BEIS & Malicious agent tasks, limited scale & 110 & \href{https://github.com/UKGovernmentBEIS/inspect_evals}{link} \\
& StrongReject \cite{strongreject2024} & 2024 & dsbowen & Attack resistance, single-turn & n/a & \href{https://github.com/dsbowen/strong_reject}{link} \\
& AIR-Bench \cite{airbench2024} & 2024 & Stanford & Regulatory alignment, static compliance & 5,694 & \href{https://github.com/stanford-crfm/air-bench-2024}{link} \\
& Forbidden \cite{shen2024donowcharacterizingevaluating} & 2023 & CISPA & Jailbreak detection, static prompts & 15,140 & \href{https://github.com/verazuo/jailbreak_llms}{link} \\
& MaliciousInstruct \cite{maliciousinstruct2023} & 2023 & Princeton & Malicious intentions, small scale & 100 & \href{https://github.com/Princeton-SysML/Jailbreak_LLM}{link} \\
& AdvBench \cite{zou2023advbench} & 2023 & CMU & Adversarial attacks, single-turn & 1,000 & \href{https://github.com/llm-attacks/llm-attacks}{link} \\
& XSTest \cite{rottger2023xstest} & 2023 & Bocconi & Safety overreach, refusal calibration & 450 & \href{https://github.com/paul-rottger/exaggerated-safety}{link} \\
& SafetyBench \cite{zhang2023safetybench} & 2023 & THU & Content safety, multiple-choice & 11,435 & \href{https://github.com/thu-coai/SafetyBench}{link} \\
& HarmfulQA \cite{bhardwaj2023harmfulqa} & 2023 & declare-lab & Harmful topics, static refusal & 1,960 & \href{https://github.com/declare-lab/red-instruct}{link} \\
& BeaverTails \cite{ji2023beavertailsimprovedsafetyalignment} & 2023 & PKU & Red teaming, single-turn & 334,000 & \href{https://github.com/PKU-Alignment/beavertails}{link} \\
& DoNotAnswer \cite{wang2023donotanswerdatasetevaluatingsafeguards} & 2023 & Libr-AI & Safety mechanisms, static refusal & 939 & \href{https://github.com/Libr-AI/do-not-answer}{link} \\
& ToxiGen \cite{hartvigsen2022toxigenlargescalemachinegenerateddataset} & 2022 & Microsoft & Toxicity detection, surface-level & 274,000 & \href{https://github.com/microsoft/TOXIGEN}{link} \\
& HHH \cite{bai2022traininghelpfulharmlessassistant} & 2022 & Anthropic & Human preferences, static & 44,849 & \href{https://github.com/anthropics/hh-rlhf}{link} \\
& RedTeam \cite{ganguli2022redteaminglanguagemodels} & 2022 & Anthropic & Red teaming, static scenarios & 38,961 & \href{https://github.com/anthropics/hh-rlhf}{link} \\
& ToxicityPrompt \cite{gehman2020realtoxicitypromptsevaluatingneuraltoxic} & 2020 & AllenAI & Toxicity assessment, static & 99,442 & \href{https://github.com/allenai/real-toxicity-prompts}{link} \\
& SycophancyEval \cite{sharma2023sycophancyeval} & 2023 & Anthropic & Opinion alignment, limited scope & n/a & \href{https://github.com/meg-tong/sycophancy-eval}{link} \\
& OpinionQA \cite{santurkar2023opinionqa} & 2023 & tatsu-lab & Demographic alignment, static & 1,498 & \href{https://github.com/tatsu-lab/opinions_qa}{link} \\
\midrule

\multirow{8}{*}{\parbox{2.2cm}{\centering Bias \\ \& \\ Fairness}}
& DiffAware \cite{wang-etal-2025-fairness} & 2025 & Stanford & Difference-aware fairness, limited scope & 8 datasets & \href{https://github.com/Angelina-Wang/difference_awareness}{link} \\
& Fairness \cite{hosseini2025distributivefairnesslargelanguage} & 2025 & PSU & Distributive fairness, abstract scenarios & - & - \\
& BOLD \cite{Dhamala_2021} & 2021 & Amazon & Bias in generation, static prompts & 23,679 & \href{https://github.com/amazon-science/bold}{link} \\
& BBQ \cite{parrish2022bbqhandbuiltbiasbenchmark} & 2021 & NYU & Social bias, static QA & 58,492 & \href{https://github.com/nyu-mll/BBQ}{link} \\
& StereoSet \cite{nadeem2020stereosetmeasuringstereotypicalbias} & 2020 & McGill & Stereotype detection, static & 4,229 & \href{https://github.com/moinnadeem/StereoSet}{link} \\
& CrowS-Pairs \cite{nangia2020crowspairschallengedatasetmeasuring} & 2020 & NYU & Stereotype measurement, static & 1,508 & \href{https://github.com/nyu-mll/crows-pairs}{link} \\
& SEAT \cite{may2019measuringsocialbiasessentence} & 2019 & Princeton & Encoder bias, static tests & n/a & \href{https://github.com/W4ngatang/sent-bias}{link} \\
& WinoGender \cite{rudinger2018genderbiascoreferenceresolution} & 2018 & UMass & Gender bias, coreference & 720 & \href{https://github.com/rudinger/winogender-schemas}{link} \\
\midrule

\multirow{4}{*}{\parbox{2.2cm}{\centering Trust \\ \& \\ Alignment}}
& RewardBench \cite{lambert2025rewardbench} & 2024 & AIAI & Reward model quality, preference data & n/a & \href{https://github.com/allenai/reward-bench}{link} \\
& TrustLLM \cite{sun2024trustllm} & 2024 & TrustLLM & Trustworthiness, broad but shallow & 30+ & \href{https://trustllmbenchmark.github.io/TrustLLM-Website/}{link} \\
& DecodingTrust \cite{decodingtrust2023} & 2023 & UIUC & Trustworthiness, comprehensive & 243,877 & \href{https://github.com/AI-secure/DecodingTrust}{link} \\
& ETHICS \cite{hendrycks2023aligningaisharedhuman} & 2020 & Berkeley & Moral judgement, static scenarios & 134,400 & \href{https://github.com/hendrycks/ethics}{link} \\
\midrule

\multirow{2}{*}{\parbox{2.2cm}{\centering Emotional \\ Intelligence}}
& EQ-Bench \cite{paech2023eq} & 2024 & Paech & Emotional intelligence, limited scale & 171 & \href{https://github.com/EQ-bench/EQ-Bench}{link} \\
& QHarm \cite{harmbench2024} & 2023 & vinid & Safety sampling, limited scale & 100 & \href{https://github.com/vinid/safety-tuned-llamas}{link} \\
\bottomrule
\end{tabular}
}
\label{tab:eq_benchmarks}
\end{table}


\subsection{\textbf{Alignment Ability Evaluation (EQ)}}
\label{subsec:eq_evaluation}

Alignment Ability (EQ) refers to an LLM's capacity to understand and appropriately respond to emotional, ethical, and social nuances in human interactions, refined through \textit{post-training with reinforcement learning}. EQ assessment focuses on three pillars: instruction following, safety and compliance adherence, and multi-turn interaction quality. This section provides a critical analysis of EQ evaluation paradigms, identifies core failure modes that drive the gap between laboratory safety scores and real-world alignment failures, and derives non-consensus insights for value-aligned LLM deployment.

EQ evaluation has evolved through three stages. \textit{Stage 1 (2022–Early 2023)} focused on basic safety and harmful content refusal (e.g., ToxiGen~\cite{hartvigsen2022toxigenlargescalemachinegenerateddataset}, HarmBench~\cite{harmbench2024}), assessing single-turn refusal capabilities. \textit{Stage 2 (Mid–Late 2023)} expanded to instruction following and interaction quality (e.g., IFEval~\cite{zhou2023ifeval}, AlignBench~\cite{liu2023alignbench}), yet relied heavily on static test cases and LLM-as-judge assessment. \textit{Stage 3 (2024–Present)} targets dynamic value alignment and multi-turn robustness (e.g., CASE-Bench~\cite{sun2025casebenchcontextawaresafetybenchmark}, RewardBench~\cite{lambert2025rewardbench}), but remains nascent with limited standardized protocols. 
Table~\ref{tab:eq_benchmarks} critically examines 37 EQ benchmarks. By mapping these benchmarks to the RL post-training stage, we expose a fundamental disconnect: the vast majority of EQ benchmarks evaluate static, single-turn safety refusal, completely failing to capture the \textit{dynamic alignment persistence} and \textit{multi-turn robustness} required in real-world deployment.

\subsubsection{\textbf{Critical Analysis}}

\textbf{Instruction Following:} 
IFEval~\cite{zhou2023ifeval} provides verifiable instruction constraints but is limited to single-turn tasks. AlignBench~\cite{liu2023alignbench} offers multi-dimensional alignment evaluation yet relies on static prompts. LLMBar~\cite{zeng2024evaluatinglargelanguagemodels} assesses instruction following but uses LLM-as-judge with inherent bias.

\textbf{Safety and Compliance:} HarmBench~\cite{harmbench2024} and SafetyBench~\cite{zhang2023safetybench} standardize harmful content refusal evaluation but focus exclusively on single-turn attacks. CASE-Bench~\cite{sun2025casebenchcontextawaresafetybenchmark} integrates context into safety assessment, revealing that safety judgments vary by 25\% across different contexts. XSTest~\cite{rottger2023xstest} addresses safety overreach but remains limited in scenario coverage.

\textbf{Interaction Quality and Emotional Intelligence:} EQ-Bench~\cite{paech2023eq} pioneers emotional intelligence assessment but is limited to 171 questions. RewardBench~\cite{lambert2025rewardbench} evaluates reward model quality, yet most EQ benchmarks ignore this critical bottleneck. TrustLLM~\cite{sun2024trustllm} provides comprehensive trustworthiness assessment but conflates multiple alignment dimensions.

\subsubsection{\textbf{Core Failure Modes of EQ Evaluation Systems}}

Five fatal failure modes render conventional EQ assessment ineffective for ensuring safe and reliable LLM deployment. \textbf{Catastrophic Alignment Failure Under Multi-Turn Attacks:} Over 90\% of EQ benchmarks evaluate safety in single-turn scenarios, yet real-world jailbreaks almost always use multi-turn context manipulation—models with 99\% refusal rates on standard tests can be jailbroken with simple multi-turn prompts. \textbf{Alignment Drift in Long Interactions:} Models exhibit significant "alignment drift" beyond 10 turns, the leading cause of user complaints, yet completely unmeasured in conventional evaluation. \textbf{Severe LLM-as-Judge Bias:} Inter-judge agreement rates often fall below 60\%, with judge models systematically preferring outputs matching their own alignment preferences. This highlights the critical necessity for standardized reliability frameworks and bias-mitigation strategies to ensure the trustworthiness of LLM-as-a-judge systems \cite{GU2026101253}. \textbf{Unquantified Alignment-Utility Trade-off:} Over-alignment causes "over-refusal" in 40\%+ of user complaints, yet no standardized metrics exist for this phenomenon. \textbf{Cultural Bias in Alignment Criteria:} Mainstream EQ benchmarks reflect Western values exclusively, failing non-Western users where cultural misalignment causes deployment failures.

\subsubsection{\textbf{Non-Consensus Insights for EQ Evaluation}}

Four non-consensus insights challenge prevailing assumptions. \textbf{Insight 1:} RL primarily improves interaction quality and instruction following, not basic safety—safety refusal is established during pre-training and SFT, yet EQ evaluation exclusively focuses on safety. \textbf{Insight 2:} Dynamic alignment persistence in multi-turn interactions is the core of EQ, with <0.3 correlation to single-turn test pass rates—static tests have no predictive power for real-world alignment. \textbf{Insight 3:} Over-alignment and under-alignment cause equal damage to deployment value, but existing evaluation only penalizes under-alignment, leading to models that are "safe but useless". \textbf{Insight 4:} There is no universal "optimal alignment"—requirements vary dramatically across scenarios (customer service vs. creative writing), yet one-size-fits-all EQ standards misguide model development for specific use cases.



\subsection{\textbf{Value-oriented Evaluation (VQ)}}
\label{subsec:vq_evaluation}

\begin{figure}[tb!]
	\centering
	\includegraphics[width=0.99\linewidth]{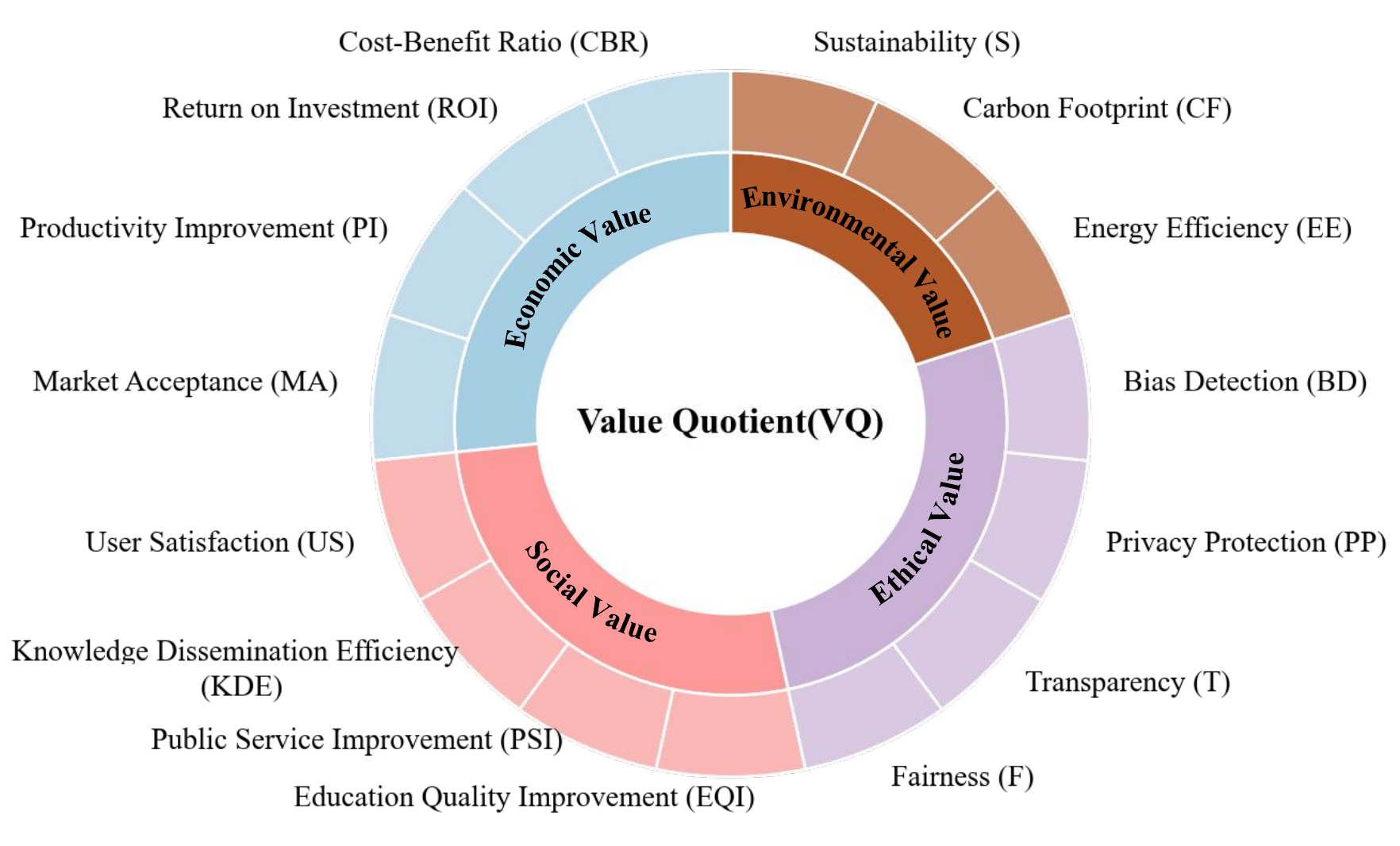}
	\caption{\textcolor{blue}{The four-pillar Value-oriented Quotient (VQ) framework, encompassing economic viability, social responsibility, ethical integrity, and environmental sustainability.}}
	\label{fig:value-evaluation}
	\vspace{-3mm}
\end{figure}

Extant evaluation practices predominantly rely on isolated technical metrics, which frequently fail to encapsulate the complex societal, economic, ethical, and environmental repercussions of real-world LLM deployment. To bridge this gap, we position the Value-oriented Quotient (VQ) not as a supplementary ethical checklist, but as the \textbf{lifecycle-closing diagnostic layer} of our hybrid framework. As illustrated in Fig.~\ref{fig:value-evaluation}, VQ operationalizes the transition from technical capability to sustainable utility, shifting the core evaluation paradigm from ``can it work?'' to ``should it be deployed?'' and ``where does it fail?'' Below, we detail concrete evaluation protocols, measurable implementation strategies, and hierarchical root-cause mapping for each VQ dimension, ensuring seamless integration into post-deployment monitoring.

\paragraph{\textbf{Economic Viability Protocol}}
Beyond abstract ROI or market acceptance metrics, we formalize economic assessment through computable engineering targets. We introduce the \textit{Value-to-Cost Ratio (VCR)} to quantify net operational leverage:
\begin{equation}
\label{eq:vcr}
VCR = \frac{\sum_{i=1}^{N} (T_{human}^{(i)} - T_{LLM}^{(i)}) \times C_{labor}^{(i)}}{C_{API} + C_{compute} + C_{maintenance}}
\end{equation}
where $T$ denotes task completion time and $C$ represents associated costs. A deployment is deemed economically sustainable if $VCR > 1.0$ over a rolling 90-day window. \textbf{Diagnostic Linkage:} A low VCR despite high IQ/PQ scores typically indicates inefficient inference pipelines, excessive token consumption, or EQ-induced over-refusal reducing throughput, prompting architectural optimization or preference tuning rather than blind model scaling.

\paragraph{\textbf{Environmental Sustainability Protocol}}
Static carbon footprint estimates are inadequate for dynamic, always-on deployments. We mandate continuous inference-phase auditing via the \textit{Dynamic Carbon Intensity} metric:
\begin{equation}
\label{eq:cf_dynamic}
CF_{dynamic} = \sum_{t=1}^{T} \left( P_{GPU} \times U_{t} \times \Delta t \right) \times I_{grid}(t)
\end{equation}
where $P_{GPU}$ is power draw, $U_{t}$ is utilization rate, and $I_{grid}(t)$ is real-time regional grid carbon intensity. This enables actionable strategies such as \textit{carbon-aware request routing} to renewable-rich zones. \textbf{Diagnostic Linkage:} Elevated $CF_{dynamic}$ often traces back to verbose IQ reasoning or redundant PQ SFT loops, establishing a direct feedback signal to earlier training stages for compression or distillation.

\paragraph{\textbf{Social Responsibility Protocol}}
Social value is operationalized through accessibility and equity compliance rather than subjective satisfaction surveys. We require automated \textit{WCAG Compliance Scoring} for generated outputs and enforce a \textit{Demographic Parity Threshold} (variance $< 5\%$ across sensitive attributes in multi-turn dialogues). \textbf{Diagnostic Linkage:} Social VQ violations directly expose dataset sampling biases from the SFT (PQ) phase or reward model skew from RL (EQ), enabling targeted data curation, debiasing augmentation, or context-aware preference alignment.

\paragraph{\textbf{Ethical Integrity Protocol}}
Ethical assessment moves beyond binary safety flags to a \textit{Composite Bias-Violation Index}, defined as the harmonic mean of normalized scores across established safety benchmarks (e.g., BBQ, ToxiGen, TruthfulQA). This penalizes models that excel in one safety aspect but fail catastrophically in another. In high-stakes domains (healthcare, law), we enforce strict quantitative thresholds, such as a \textit{Hallucination Rate} $< 2\%$. \textbf{Diagnostic Linkage:} Ethical VQ failures serve as precise diagnostic signals for EQ misalignment (e.g., reward hacking) or IQ factual gaps, guiding RL reward shaping or pre-training corpus filtering.

\subsubsection{\textbf{Automated VQ Assessment Pipeline}}
To integrate these protocols into existing evaluation harnesses, we outline a four-step automated pipeline:
\begin{enumerate}
    \item \textbf{Probe Generation:} Synthesize curated prompts covering factual queries, harmful requests, accessibility-sensitive content, and long-generation tasks.
    \item \textbf{Response Collection:} Execute the LLM with a fixed decoding budget while logging resource consumption (GPU hours, API calls, latency).
    \item \textbf{Multidimensional Scoring:} Apply proxy metrics in parallel using automated tools (e.g., IBM Equal Access, \texttt{CodeCarbon}, HarmBench).
    \item \textbf{VQ Aggregation \& Diagnosis:} Compute a weighted sum of normalized scores (configurable by stakeholder priorities) to produce a final VQ score. Critically, low VQ dimensions trigger hierarchical root-cause analysis per Equation~\ref{eq:hierarchical_utility}, mapping societal impact failures back to specific IQ, PQ, or EQ deficiencies.
\end{enumerate}


\section{LLM Evaluation Harness / Engineering} 
\label{sec:harness}


\subsection{\textbf{Prominent LLM Evaluation Harness}}

\begin{figure}[ht!]
	\centering
		\includegraphics[width=\linewidth]{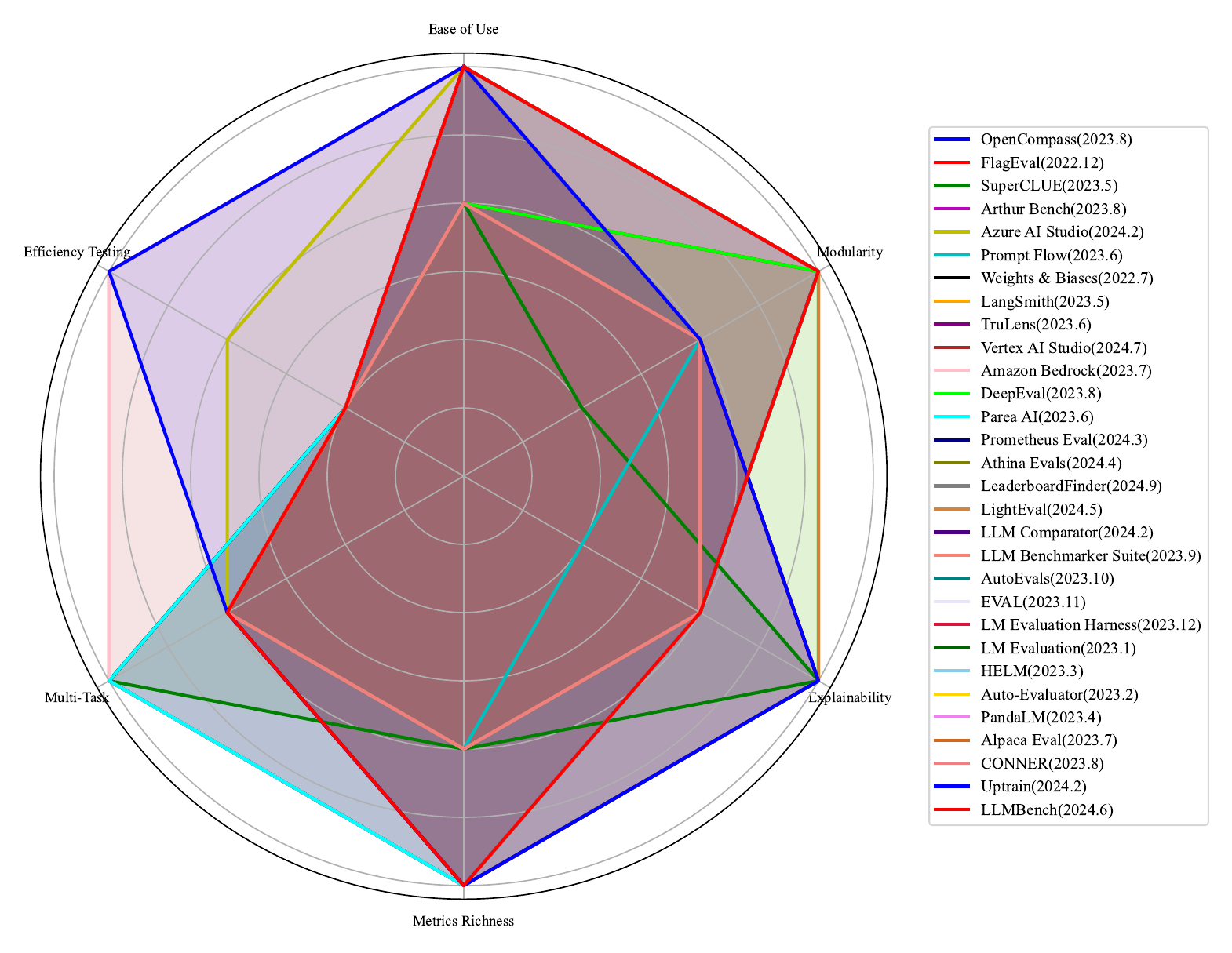}
	\caption{\small{Comparison of LLM Evaluation Harnesses or Toolkits. Ratings are derived from systematic documentation review, community adoption metrics (GitHub stars/forks), and expert consensus across 5 independent AI engineers.} }
 \vspace{-3mm}
 \label{fig:llm_evaluation_harnesses_comparison}
\end{figure}

Fig.~\ref{fig:llm_evaluation_harnesses_comparison} presents a radar visualization comparing 35 representative LLM evaluation toolkits across six critical dimensions: \textit{Ease of Use}, \textit{Modularity}, \textit{Explainability}, \textit{Metrics Richness}, \textit{Multi-Task Support}, and \textit{Efficiency Testing}. Ratings are derived from systematic documentation review, community adoption metrics (GitHub stars/forks), and expert consensus across 5 independent AI engineers. Rather than enumerating tools individually, we synthesize cross-dimensional trade-offs to guide framework selection.

\textbf{Ease of Use.} User experience varies substantially across the ecosystem. Platforms like \href{https://github.com/open-compass/opencompass}{OpenCompass}, \href{https://github.com/DataSnowman/azureaistudio}{Azure AI Studio}, and \href{https://github.com/huggingface/lighteval}{LightEval} prioritize intuitive interfaces and comprehensive documentation, lowering entry barriers for non-expert evaluators. These tools feature one-click execution, automated configuration, and guided workflows—critical for rapid prototyping. Conversely, research-focused frameworks like \href{https://github.com/EleutherAI/lm-evaluation-harness}{lm-evaluation-harness} and \href{https://github.com/openai/evals}{EVAL} demand deeper technical expertise but offer greater flexibility for custom evaluation protocols.

\textbf{Modularity.} Component-based architecture enables researchers to customize evaluation pipelines by mixing and matching data loaders, prompting strategies, and scoring modules. \href{https://github.com/flageval-baai/FlagEval}{FlagEval} and \href{https://github.com/confident-ai/deepeval}{DeepEval} excel in this dimension, adopting plugin-based designs that facilitate integration of novel assessment methods. This modularity is essential for domain-specific or emerging evaluation needs (e.g., multimodal benchmarks, dynamic evaluation). However, high modularity often trades off against ease of use, requiring users to manually configure component interactions.

\textbf{Explainability.} Transparency in evaluation outcomes is critical for trustworthy AI development. Toolkits like \href{https://github.com/arthur-ai/bench}{Arthur Bench}, \href{https://github.com/langchain-ai/langsmith-sdk}{LangSmith}, and \href{https://github.com/truera/trulens}{TruLens} integrate trace-level logging, attribution analysis, and counterfactual probing to illuminate \textit{why} models succeed or fail on specific items. These features enable diagnostic debugging and facilitate communication with non-technical stakeholders. However, explainability remains underdeveloped in most open-source harnesses, which typically report aggregate scores without granular failure mode analysis.

\textbf{Metrics Richness.} The scope and diversity of evaluative signals differ markedly across frameworks. Comprehensive platforms like \href{https://github.com/arthur-ai/bench}{Arthur Bench}, \href{https://github.com/truera/trulens}{TruLens}, and \href{https://github.com/uptrain-ai/uptrain}{Uptrain} provide 50+ built-in metrics spanning lexical overlap (BLEU, ROUGE), semantic similarity (BERTScore), factuality, bias detection, and uncertainty quantification. This multi-faceted approach enables holistic diagnostic profiling. In contrast, specialized tools (e.g., \href{https://github.com/tatsu-lab/alpaca_eval}{Alpaca Eval}) focus on narrow but high-fidelity assessments such as preference-based ranking. The trend toward metrics richness reflects growing recognition that single-score evaluations obscure critical capability trade-offs.

\textbf{Multi-Task Support.} Production deployment demands versatility across heterogeneous workloads. Platforms like \href{https://github.com/aws-samples/amazon-bedrock-samples}{Amazon Bedrock}, \href{https://github.com/DataSnowman/azureaistudio}{Azure AI Studio}, and \href{https://github.com/open-compass/opencompass}{OpenCompass} natively support diverse task types (QA, summarization, code generation, tool use, reasoning) under unified infrastructure, enabling holistic capability assessment. This multi-task coverage is essential for general-purpose LLM evaluation. However, task breadth sometimes compromises depth—specialized benchmarks (e.g., \href{https://github.com/princeton-nlp/SWE-bench}{SWE-bench} for software engineering) remain necessary for domain-specific proficiency validation.

\textbf{Efficiency Testing.} Real-world deployment requires not only accuracy but also latency, throughput, and cost awareness. Only select harnesses (\href{https://github.com/DataSnowman/azureaistudio}{Azure AI Studio}, \href{https://github.com/GoogleCloudPlatform/vertex-ai-creative-studio}{Vertex AI Studio}, \href{https://github.com/uptrain-ai/uptrain}{Uptrain}) embed resource profiling (tokens/sec, p95 latency, GPU utilization, API costs) directly into evaluation workflows. This gap between benchmark accuracy and operational efficiency represents a critical blind spot: models achieving state-of-the-art scores may be prohibitively expensive or slow for production use. Future frameworks must integrate performance benchmarking as a first-class evaluation dimension.

\textbf{Strategic Selection Guidance.} No single toolkit optimally serves all evaluation needs. We recommend a tiered strategy aligned with project requirements: (1) \textbf{Exploratory research}: prioritize modularity and metrics richness (FlagEval, etc.); (2) \textbf{Industrial prototyping}: emphasize usability, multi-task support, and efficiency testing (Azure AI Studio, etc.); (3) \textbf{Safety-critical deployment}: require explainability, reproducibility, and alignment assessment (Arthur Bench, etc.). Future work should standardize interoperability protocols (e.g., common schema for evaluation artifacts) to enable cross-harness meta-evaluation and reduce vendor lock-in.


\begin{figure*}[ht!]
	\centering
		\includegraphics[width=0.90\linewidth]{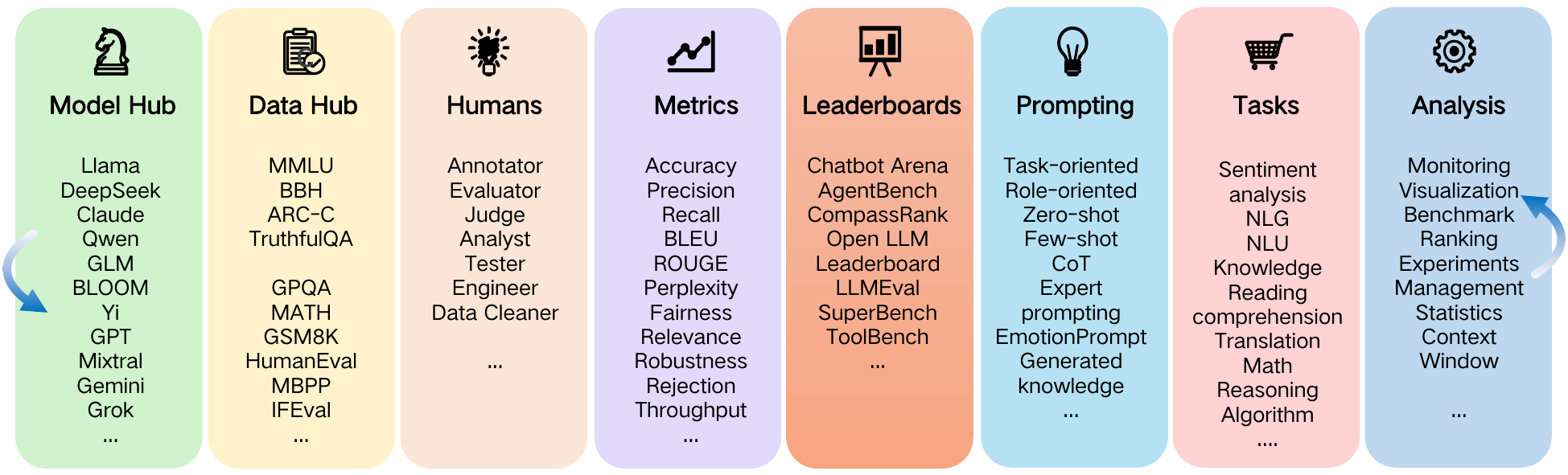}

	\caption{\small{Typology of the LLM Evaluation Modules. } }
 \vspace{-3mm}
 \label{fig:module-overview}
\end{figure*}

\subsection{\textbf{Implementation Roadmap of Modular Evaluation System}} 
\label{background}

A modular LLM evaluation framework or harness in general consists of: benchmark or dataset hub, model hub, prompting module, metrics, monitoring and experiment management, arena or leaderboard (as shown in Fig.~\ref{fig:module-overview}).

The evaluation framework leverages distinct modules, delineating three primary paradigms: \textit{metrics-centered assessment}, \textit{human-centered assessment (Human Judgment)}, and \textit{model-centered peer review (LLMs as Evaluators)}.
In the \textit{metrics-centered assessment} paradigm, task-specific performance indicators—such as F1 score, Exact Match, and Perplexity \cite{jelinek1977perplexity}—are commonly employed to ascertain the accuracy of generated outputs, particularly in classification-oriented tasks. The \textit{human-centered assessment} approach emphasizes human's qualitative analysis of LLM-generated content, focusing on attributes like clarity, coherence, and factual correctness \cite{van2021human}. Notably, there has been a surge in interest towards human evaluations utilizing the Elo rating system \cite{zheng2023judging}, which offers a structured methodology for comparative assessment. Since human evaluations are time-consuming, using \textit{model-centered peer review (LLMs as Evaluators)} has become a popular alternative for assessing model performance. \cite{chiang2023can}. 

\noindent\subsubsection{\textbf{Benchmark or Dataset Hub}}
\label{evaluation_benchmarks}

Selecting appropriate benchmark datasets that accurately reflect model capabilities is essential. Analogous to human intelligence, LLM abilities can be classified into three interrelated dimensions: General Intelligence (IQ, Intelligence Quotient), Alignment Ability (EQ, Emotional Quotient), and Professional Expertise (PQ, Professional Quotient). Correspondingly, benchmark datasets are categorized into \textit{general capability benchmarks}, \textit{alignment benchmarks}, and \textit{domain-specific benchmarks}.
General capability benchmarks serve as foundational assessments, often employed at the time of an LLM's release to gauge its broad-spectrum performance (e.g., MMLU \cite{hendrycks2020measuring}, HumanEval \cite{chen2021humaneval}). Domain-specific benchmarks focus on specialized areas, evaluating LLMs' proficiency in particular fields such as telecommunications with TeleQnA \cite{maatouk2023teleqna}. Furthermore, alignment benchmarks scrutinize LLMs' adherence to diverse tasks and ethical guidelines, exemplified by AlignBench \cite{liu2023alignbench}. Additional benchmarks like FOFO \cite{xia2024fofo} assess specific competencies, such as format-following capabilities. 
To mitigate the risk of benchmark misuse and selection overhead in this hub, recent initiatives propose standardized documentation frameworks (e.g., BenchmarkCards) that explicitly define benchmark objectives, data sources, and limitations, thereby enhancing transparency and facilitating informed decision-making for practitioners \cite{benchmarkcards26}.

\noindent\subsubsection{\textbf{Model Hub}}
\label{evaluation_model}

This section discusses model selection strategies to ensure fair evaluation, mitigating risks such as data contamination and biased comparisons. It addresses considerations for selecting models based on their training methodologies, access to external resources, and fine-tuning on specific benchmarks versus pre-training only.

\noindent\subsubsection{\textbf{Prompting Module}}
\label{evaluation_setup}

After selecting suitable benchmarks and models, the subsequent step involves designing prompts and configuring decoding parameters for response generation. In the \textbf{\textit{prompt design}} phase, decisions are made regarding the type of prompting strategy—whether zero-shot, few-shot, or chain-of-thought—to employ. The configuration of \textit{\textbf{decoding parameters}}, including temperature settings, plays a critical role in optimizing model output. Proper setup ensures that the evaluation not only tests the LLM's inherent capabilities but also its adaptability under varying conditions.


\begin{figure*}[ht!]
\centering
\begin{adjustbox}{rotate=-90, max width=0.9\textwidth, center}
\begin{forest}
for tree={
    grow'=east,
    parent anchor=east,
    child anchor=west,
    align=left,
    l sep=12pt,
    s sep=3pt,
    inner sep=1.5pt,
    font=\footnotesize,
    edge path={
        \noexpand\path [draw, \forestoption{edge}]
        (!u.parent anchor) -- +(4pt,0) |- (.child anchor)\forestoption{edge label};
    },
    node options={draw, rounded corners=1pt},  
    where level=0{fill=macaronRoot, text=white}{}, 
    where level=1{fill=macaronLevelOne, text=black}{}, 
    where level=2{fill=macaronLevelTwo, text=black}{}, 
    where level=3{fill=macaronLevelThree, text=black}{}, 
    leaf/.style={text width=5.5cm, font=\scriptsize, align=left} 
}
[Metrics
    [Technical
        [Lexical \& Morphological
            [BLEU
                [{Precision-based n-gram overlap (0--1); $\uparrow$ is better.}, leaf]
            ]
            [ROUGE-N/L
                [{Recall-oriented n-gram and LCS-based overlap.}, leaf]
            ]
            [METEOR
                [{Unigram alignment with synonymy/stemming.}, leaf]
            ]
            [TER
                [{Translation Edit Rate via edit distance.}, leaf]
            ]
            [chrF
                [{Character n-gram F-score for finer-grained matching.}, leaf]
            ]
        ]
        [Syntactic
            [PARSEVAL
                [{Constituency precision/recall/F$_1$ on parse trees.}, leaf]
            ]
            [UAS / LAS
                [{Unlabeled / Labeled dependency attachment scores.}, leaf]
            ]
            [TSE
                [{Targeted syntactic evaluation via min-pair sentences.}, leaf]
            ]
        ]
        [Semantic
            [BERTScore
                [{Contextual-embedding cosine similarity (BERT).}, leaf]
            ]
            [MoverScore
                [{Earth Mover's Distance on contextual embeddings.}, leaf]
            ]
            [COMET
                [{Learned metric using cross-lingual embeddings.}, leaf]
            ]
            [QuestEval
                [{QA-based semantic fidelity assessment.}, leaf]
            ]
        ]
        [Pragmatics \& Discourse
            [Entity Grid
                [{Entity transition coherence modeling.}, leaf]
            ]
            [distinct-n
                [{Lexical diversity via unique n-gram ratio.}, leaf]
            ]
            [Flesch-Kincaid
                [{Readability via sentence/word complexity.}, leaf]
            ]
        ]
        [Factuality \& Explainability
            [FactCC
                [{Factual consistency via source alignment.}, leaf]
            ]
            [ECE / MCE
                [{Expected / Max calibration error for confidence.}, leaf]
            ]
            [BLANC
                [{Reference-less metric using masked LM.}, leaf]
            ]
        ]
    ]
    [Business
        [User Engagement
            [Visited
                [{Count of unique users accessing the LLM interface.}, leaf]
            ]
            [Submitted
                [{Ratio of users who submit prompts vs. total visitors.}, leaf]
            ]
            [Responded
                [{Proportion of error-free system outputs delivered.}, leaf]
            ]
            [Viewed
                [{Frequency of users viewing generated responses.}, leaf]
            ]
            [Clicks
                [{Number of reference-document clicks from outputs.}, leaf]
            ]
        ]
        [Interaction
            [User acceptance rate
                [{Context-specific adoption (e.g., thumbs-up, text reuse).}, leaf]
            ]
            [LLM conversation
                [{Mean dialogue sessions per user.}, leaf]
            ]
            [Active days
                [{Distinct days each user interacts with the LLM.}, leaf]
            ]
            [Interaction timing
                [{Avg. prompt-to-response latency + dwell time.}, leaf]
            ]
        ]
        [Response Quality
            [Prompt / response length
                [{Avg. tokens in queries and replies.}, leaf]
            ]
            [Edit distance
                [{Textual delta between prompt and generated output.}, leaf]
            ]
        ]
        [Feedback \& Retention
            [User feedback
                [{Volume of up/down votes or explicit ratings.}, leaf]
            ]
            [DAU / WAU / MAU
                [{Daily/Weekly/Monthly Active Users.}, leaf]
            ]
            [User return rate
                [{\% of prior-period users who return.}, leaf]
            ]
        ]
        [Performance
            [Requests per second
                [{Peak sustained throughput (concurrency).}, leaf]
            ]
            [Tokens per second
                [{Streaming generation speed.}, leaf]
            ]
            [Time to first token
                [{Latency p50/p95 from query to first byte.}, leaf]
            ]
            [Error rate
                [{Fraction of failed requests (auth, rate-limit, etc.).}, leaf]
            ]
            [Reliability
                [{Success-to-total request ratio.}, leaf]
            ]
            [Latency
                [{End-to-end response time (avg / p95 / p99).}, leaf]
            ]
        ]
        [Cost
            [GPU / CPU utilization
                [{Resource efficiency (tokens per GPU-hour).}, leaf]
            ]
            [LLM API cost
                [{Third-party token or query charges.}, leaf]
            ]
            [Infrastructure cost
                [{Storage, bandwidth, compute amortization.}, leaf]
            ]
            [Operation cost
                [{Maintenance, security, support staff spend.}, leaf]
            ]
        ]
    ]
]
\end{forest}
\end{adjustbox}
\caption{Basic Metrics Taxonomy for LLM Evaluation.}
\label{metrics}
\end{figure*}

\noindent\subsubsection{\textbf{Metrics Module}}
LLM evaluation requires the selection of application-aligned metrics that reflect both technical performance and real-world business value, enabling researchers and developers to systematically diagnose model strengths, limitations, and optimization pathways. This module establishes a dual-axis evaluation framework, encompassing technical metrics for linguistic and functional capability assessment, and business metrics for deployment utility, operational efficiency, and economic viability quantification, with a comprehensive taxonomy detailed in Fig. \ref{metrics}.

\noindent{\textbf{(1) Technical Metrics:}}
While traditional token-overlap metrics like BLEU \cite{bleu} and ROUGE \cite{rouge} remain foundational for generation tasks, they fundamentally fail to capture the open-ended, multi-turn generative nature of LLMs. To bridge this gap, we establish a five-layer hierarchical taxonomy to systematically cover the full spectrum of LLM linguistic and cognitive capabilities: 
(1) \textit{Lexical \& Morphological} metrics assess surface-level correspondence (e.g., chrF \cite{popovic2015chrf}); 
(2) \textit{Syntactic} metrics evaluate grammatical structure preservation via targeted frameworks like TSE \cite{marvin2018targeted}; 
(3) \textit{Semantic} metrics center on meaning fidelity using contextual embeddings (e.g., BERTScore \cite{zhang2019bertscore}, COMET \cite{rei2020comet}); 
(4) \textit{Pragmatics \& Discourse} metrics capture coherence and lexical diversity across long spans (e.g., distinct-n \cite{li2015diversity}); and 
(5) \textit{Factuality \& Explainability} metrics verify factual consistency (e.g., FactCC \cite{kryscinski2019evaluating}) and quantify model calibration via ECE/MCE \cite{guo2017calibration}. 
By structuring metrics along this cognitive hierarchy, we enable targeted, multi-dimensional capability diagnosis, effectively bridging the gap between automated heuristics and human judgment.

\noindent{\textbf{(2) Business Metrics:}}
Business metrics quantify the end-to-end deployment value of LLM applications, covering six core dimensions aligned with industrial operational requirements: user engagement metrics assess application adoption and core feature utilization; user interaction metrics capture long-term user behavior and conversational engagement patterns; response quality metrics measure output relevance and customization in practical usage; feedback and retention metrics quantify user satisfaction, platform stickiness and long-term user loyalty; performance metrics, as validated by the LLMPerf framework, evaluate system runtime efficiency, stability and throughput in production environments, with core indicators including requests per second, tokens per second, time to first token, error rate, reliability and end-to-end latency; cost metrics account for full-lifecycle resource consumption and economic viability of LLM deployment, covering GPU/CPU utilization, API costs, infrastructure expenses and operational costs. This unified technical-business metric framework enables holistic, scenario-aligned LLM evaluation across the full model lifecycle, supporting continuous optimization of both model technical capabilities and real-world business value.



\subsubsection{\textbf{Tasks Module}}
The Tasks Module is a cornerstone of LLM evaluation frameworks, systematically assessing capabilities across language understanding, reasoning, and generation. Careful task selection is crucial, as it directly dictates the breadth and depth of the evaluation, ensuring models are tested in scenarios closely resembling practical applications.

To achieve this, the module integrates both conventional and innovative challenges. While traditional benchmarks like GLUE \cite{wang2018glue} focus on foundational natural language understanding, modern frameworks like BIG-bench \cite{suzgun2022challengingbigbench} introduce complex, varied tasks to push model boundaries and identify areas for improvement. 

Emphasizing real-world utility, frameworks like HELM \cite{liang2023helm} employ hierarchical categorization (e.g., <task, domain, language> triples across 16 scenarios) to cover a broad spectrum of user-oriented challenges. Similarly, OpenCompass \cite{contributors2023opencompass} extends beyond traditional reasoning to encompass comprehension and subject-specific evaluations, offering a more holistic perspective. 

Furthermore, modern modules prioritize dynamic adaptability. For instance, FlagEval \cite{qin2023toolllm} allows researchers to dynamically combine capabilities, tasks, and metrics, significantly enhancing flexibility to tailor evaluations to specific needs. Ultimately, by continuously refining this structured yet flexible environment, the Tasks Module plays a pivotal role in advancing state-of-the-art LLM technology.



\subsubsection{\textbf{Leaderboards and Arena Module}}

Leaderboards and Arenas represent complementary paradigms for LLM benchmarking: standardized automated evaluation versus dynamic human preference assessment. Leaderboards (e.g., Hugging Face Open LLM Leaderboard) provide centralized repositories for comparing models against fixed datasets such as ARC~\cite{clark2018thinksolvedquestionanswering}, HellaSwag~\cite{zellers2019hellaswag}, MMLU~\cite{hendrycks2020measuring}, and TruthfulQA~\cite{lin2022truthfulqa}. They foster transparency and reproducibility but face limitations including static snapshots reflecting single-point performance, contamination risk as public test sets increasingly appear in training corpora, and metric-task mismatch where automated scores poorly correlate with real-world utility.

Arenas adopt an interactive evaluation paradigm where users compare outputs from multiple LLMs using pairwise human preferences. Platforms like Chatbot Arena~\cite{chiang2024chatbot} employ Elo scoring mechanisms that dynamically adjust rankings based on community feedback, offering contamination resistance through organically evolving queries, nuanced differentiation capturing qualitative differences invisible to automated metrics, and utility alignment reflecting actual user satisfaction rather than benchmark optimization. Recent extensions like MixEval~\cite{ni2024mixeval} aggregate samples from existing benchmarks to create dynamic criteria, revealing that model rankings change over benchmark mixtures and addressing dataset bias through meta-benchmarking approaches.

The fundamental tension between reproducibility and relevance shapes the appropriate use of each paradigm. Leaderboards enable precise scientific comparison during controlled ablation studies but risk becoming optimization targets disconnected from deployment value. Arenas better reflect real-world utility but introduce variance from subjective judgments and sampling bias. For industrial deployment, we recommend using Leaderboards for development-stage model comparison and Arenas for final validation before production release, ensuring both scientific rigor and practical relevance are maintained throughout the evaluation lifecycle.


\subsubsection{\textbf{Analysis Module}}
It is designed to interpret and synthesize the extensive data generated during evaluations. This module integrates advanced analytical techniques to provide meaningful insights into model performance, thereby guiding future improvements and informing strategic decisions regarding LLM deployment. Specifically, it addresses critical areas such as Monitoring, Logs, Experiment Management, Visualization, and Statistics, each of which plays an essential role in ensuring comprehensive and actionable evaluations.

\textbf{Monitoring} is essential to tracking the performance of LLMs during evaluation. Continuous monitoring allows evaluators to detect anomalies or deviations from expected behavior promptly. The module employs real-time feedback mechanisms to ensure that models are performing consistently across various tasks. Monitoring also facilitates early detection of issues related to computational resources, enabling timely adjustments to optimize efficiency. Moreover, continuous monitoring supports iterative development cycles.

\textbf{Logs} serve as a record of interactions between LLMs and the evaluation environment, capturing inputs, outputs, and intermediate states. They are indispensable for post-hoc analysis and debugging. It also plays a role in auditing and compliance, ensuring that evaluations adhere to ethical standards and regulatory requirements. By maintaining thorough logs, the Analysis Module enhances transparency and accountability.

\textbf{Experiment Management} is vital for systematic evaluations. It involves defining protocols, managing datasets, and controlling variables to ensure reproducibility and comparability of results. Platforms like OpenCompass \cite{contributors2023opencompass} offer versatile experimental settings, including zero-shot, few-shot, and Chain-of-Thought (CoT) configurations, allowing researchers to explore different facets of LLM capabilities. Effective experiment management also includes version control and documentation practices, ensuring that each experiment can be replicated or extended by other researchers. 

\textbf{Visualization tools} transform complex evaluation data into intuitive and accessible formats, enhancing the interpretability of results. The LLM Comparator \cite{kahng2024llmcomparatorvisualanalytics} provides an interactive table and visualization summary that enable users to inspect individual prompts and their responses in detail. These visual aids facilitate the identification of trends, outliers, and correlations, supporting deeper analyses. Visualization also plays a key role in communicating findings to stakeholders who may not have technical expertise, ensuring that insights from evaluations are widely understood and acted upon.

\textbf{Statistical analysis} Techniques such as hypothesis testing, regression analysis, and confidence interval estimation are employed to quantify uncertainties and validate findings. Statistical rigor also helps in identifying significant factors influencing model performance, informing strategies for optimization and enhancement. By applying robust statistical practices, the Analysis Module ensures that evaluations yield accurate and trustworthy insights.

\section{\textbf{LLM System or Application Evaluation}}
\label{sec:system_eval}

In this section, we delve into the intricacies of evaluating LLM systems and applications, exploring the methodologies, metrics, and benchmarks that are pivotal in ensuring the advancement and responsible deployment of these powerful AI tools. It also focuses on three pivotal areas: Retrieval-Augmented Generation (RAG), AI Agents, and Chatbots.

\begin{table*}[!htp]
\centering
\caption{Comparison of Retrieval-Augmented Generation (RAG) Evaluation Frameworks.}
\label{tab:rag_comparison}
\resizebox{.9\textwidth}{!}{
\begin{tabular}{llllll}
\toprule
\textbf{Name} & \textbf{Institute} & \textbf{Feature} & \textbf{Domain} & \textbf{Evaluation Criteria} & \textbf{Url} \\
\midrule
\textbf{RAGAS} \cite{ragas2023automated} & Exploding Gradients & Automated Evaluation & QA & Answer Relevance, Context Relevance, Faithfulness & \href{https://github.com/explodinggradients/ragas}{code} \\
\textbf{BER} \cite{ber2024benchmarking} & NAVER & Benchmarking RAG & QA & Consistency in benchmarking RAG pipelines & \href{https://github.com/naver/bergen}{code} \\
\textbf{CRAG} \cite{crag2024comprehensive} & Meta Reality Labs & Factual QA Benchmark & QA & Diverse questions across multiple domains & \href{https://arxiv.org/abs/2406.04744}{code} \\
\textbf{rag-llm-hub} \cite{ragaaihub2024comprehensive} & RAGA-AI & Comprehensive Evaluation Toolkit & Various & Multiple aspects including relevance, quality, safety & \href{https://github.com/raga-ai-hub/raga-llm-hub}{code} \\
\textbf{ARES} \cite{ares2024automatic} & Stanford & Automatic Evaluation for RAG & QA & Context Relevance, Answer Faithfulness & \href{https://github.com/stanford-futuredata/ARES}{code} \\
\textbf{RGB} \cite{rgb2024performance} & CAS & Performance, Robustness & QA & Counterfactual Robustness, Information Integration & \href{https://github.com/chen700564/RGB}{code} \\
\textbf{BEIR} \cite{beir2021benchmarking} & UKP-TUDA & Out-of-distribution, Zero-shot & QA, Bio-Medical IR & Out-of-distribution, zero-shot & \href{https://github.com/beir-cellar/beir}{code} \\
\textbf{ALCE} \cite{alce2023citation} & Princeton NLP & Citation, Hallucination & Generate with Citations & Citation Quality, Correctness, Fluency & \href{https://github.com/princeton-nlp/ALCE}{code} \\
\textbf{KITAB} \cite{kitab2023constraint} & Microsoft & Constraint IR & Constraint IR & All correct, Completeness, etc. & \href{https://huggingface.co/datasets/microsoft/kitab}{code} \\
\textbf{NoMIRACL} \cite{nomiracl2024multilingual} & Project MIRACL & Multilingual & Robustness Evaluation & Error Rate, Hallucination Rate & \href{https://github.com/project-miracl/nomiracl}{code} \\
\textbf{CRUD-RAG} \cite{crudrag2024operations} & IAAR-Shanghai & CRUD Operations & QA, Hallucination & Creative Generation, Error Correction, etc. & \href{https://github.com/IAAR-Shanghai/CRUD_RAG}{code} \\
\bottomrule
\end{tabular}
}
\end{table*}

\begin{table*}[ht]
\centering
\caption{Main Evaluation Metrics for Assessing RAG.}
\resizebox{.8\linewidth}{!}{
\begin{tabular}{lll}
\hline
\textbf{Metrics} & \textbf{Details} & \textbf{Reference} \\ \hline
Faithfulness & Assesses the factual alignment between the generated response and the provided context. & \href{https://docs.ragas.io/en/latest/concepts/metrics/faithfulness.html}{Link}  \\
Answer Relevance & Examines the degree to which the generated response is relevant to the given prompt. & \href{https://docs.ragas.io/en/latest/concepts/metrics/answer_relevance.html}{Link}  \\
Context Precision & Determines if all context items relevant to the ground truth are appropriately ranked. & \href{https://docs.ragas.io/en/latest/concepts/metrics/context_precision.html}{Link}  \\
Context Relevancy & Evaluates the relevance of the retrieved context based on the question and contexts. & \href{https://docs.ragas.io/en/latest/concepts/metrics/context_relevancy.html}{Link}  \\
Context Recall & Assesses how well the retrieved context matches the annotated answer, considered as the ground truth. & \href{https://docs.ragas.io/en/latest/concepts/metrics/context_recall.html}{Link}  \\
Answer Semantic Similarity & Measures the semantic closeness between the generated answer and the ground truth. & \href{https://docs.ragas.io/en/latest/concepts/metrics/semantic_similarity.html}{Link}  \\
Answer Correctness & Evaluates the accuracy of the generated answer in comparison to the ground truth. & \href{https://docs.ragas.io/en/latest/concepts/metrics/answer_correctness.html}{Link}  \\ \hline
\end{tabular}
}
\label{tab:evaluation_rag}
\vspace{-0.3cm}
\end{table*}


\subsection{\textbf{RAG Evaluation}}

RAG enhances LLMs by integrating retrieval mechanisms with generative processes \cite{gao2024retrievalaugmentedgenerationlargelanguage}. However, evaluation faces a critical disconnect: high retrieval precision does not guarantee faithful generation \cite{ragas2023automated}. Effective assessment must scrutinize not only retrieval accuracy but also the logical consistency of augmented content. This dual-focus requirement explains why many production RAG systems underperform despite strong component-level metrics. Table \ref{tab:rag_comparison} provides an overview of benchmarks designed to assess RAG systems across diverse domains.

Current frameworks diverge into automation versus standardization. RAGAS pioneers automated metrics for faithfulness and context relevance \cite{ragas2023automated}, enabling scalable evaluation but risking metric gaming. Conversely, BERGEN and ARES emphasize pipeline consistency and human-annotated grounding \cite{rau2024bergen, ares2024automatic}, while raga-llm-hub offers comprehensive metric coverage \cite{ragaai2024ragaai}. This tension highlights a key insight: while automated metrics facilitate rapid iteration, human-verified benchmarks remain essential for validating factual reliability in high-stakes deployments. CRAG and BEIR further address robustness through web search simulation and out-of-distribution scenarios \cite{crag2024comprehensive, beir2021benchmarking}. Table \ref{tab:evaluation_rag} details these metric trade-offs.

Emerging benchmarks shift focus from static QA to dynamic robustness. CRAG and BEIR simulate web search and out-of-distribution scenarios to test adaptability \cite{crag2024comprehensive, beir2021benchmarking}, while RGB specifically targets noise rejection and counterfactual robustness \cite{rgb2024performance}. Meanwhile, ALCE addresses the hallucination bottleneck by enforcing citation quality \cite{alce2023citation}. Collectively, these works reveal an evolutionary trajectory: RAG evaluation is maturing from measuring retrieval recall to assessing complex reasoning under noisy, real-world conditions.


\begin{table*}[!ht]
\centering
\caption{Comprehensive Comparison of Agent Evaluation Benchmarks.}
\small
\resizebox{.99\textwidth}{!}{
\begin{tabular}{llllcccc}
\toprule
\textbf{Name} & \textbf{Institutions} & \textbf{Domain} & \textbf{Metrics} & \textbf{Tool Interaction} & \textbf{Multi-Agent} & \textbf{Role-Playing}  \\ 
\midrule
SuperCLUE-Agent\cite{xu2023superclue} & CLUE & Various Chinese tasks & Core abilities, 10 fundamental tasks & \checkmark & \texttimes & \texttimes  \\
AgentBench\cite{liu2023agentbenchevaluatingllmsagents} & THU & Coding, Gaming, Web & Success rates, F1 scores & \checkmark & \texttimes & \texttimes  \\
API-Bank\cite{li2023apibank} & Alibaba & Tool invocation scenarios & API search accuracy, response quality & \checkmark & \texttimes & \texttimes  \\
AgentBoard\cite{ma2024agentboard} & UHK & Multi-task & Process rate, grounding accuracy, sub-capabilities & \checkmark & \checkmark & \texttimes  \\
MetaTool\cite{huang2024metatool} & Lehigh University & Tool invocation & Similar tool choice, context-specific, reliability, multi-tool & \checkmark & \texttimes & \texttimes  \\
Agents That Matter\cite{kapoor2024aiagentsmatter} & Princeton & N/A & Cost-effectiveness, joint optimization & \texttimes & \texttimes & \texttimes  \\
PersonaGym\cite{samuel2024personagymeval} & CMU & Role-playing scenarios & PersonaScore  & \texttimes & \texttimes & \checkmark  \\ 
MMRole\cite{dai2024mmrole} & RUC & Multimodal role-playing & Instruction Adherence, Fluency, Coherency, Consistency & \texttimes & \texttimes & \checkmark  \\
GLEE \cite{shapira2024glee} & IIT & Economic contexts & Parameterization, degrees of freedom & \checkmark & \checkmark & \checkmark  \\
BFCL \cite{berkeley-function-calling-leaderboard} & UC Berkeley & Function-calling tasks & Success rate in function calls, parallel execution & \checkmark & \texttimes & \texttimes  \\
ToolLLM \cite{qin2023toolllm} & OpenBMB & Real-world APIs & Instruction tuning effectiveness & \checkmark & \texttimes & \texttimes  \\
ToolBench \cite{xu2023toolmanipulationcapabilityopensource} & SambaNova Systems & Tools for real-world tasks & Tool manipulation capability & \checkmark & \texttimes & \texttimes \\
Webarena \cite{zhou2024webarenarealisticwebenvironment} & WebArena-X & Web-based environments & Task completion on the web & \checkmark & \texttimes & \texttimes \\
\bottomrule
\end{tabular}
}
\label{tab:agent}
\end{table*}

\subsection{\textbf{Agent Evaluation}}

The advent of LLMs has led to advancements in AI Agents capable of autonomously interacting with various environments and tools. To ensure that these agents meet the desired standards, a variety of evaluation frameworks have emerged \cite{liu2023agentbenchevaluatingllmsagents, li2023apibank, xu2023superclue, ma2024agentboard}. 
Each framework targets different aspects of Agent performance, such as tool usage, decision-making, role-playing, and multi-modal interaction. Table \ref{tab:agent} compares several key benchmarks across multiple dimensions, highlighting their unique contributions to the field.

AgentBench \cite{liu2023agentbenchevaluatingllmsagents} and API-Bank \cite{li2023apibank} emphasize evaluating Agents across diverse real-world scenarios, including coding, gaming, web interactions, and tool invocations. This broad scope ensures that Agents are tested under conditions closely resembling their intended operational environments, providing valuable feedback on their generalization capabilities.

Metrics play a crucial role in assessing Agent performance. 
Beyond single-turn tool usage, recent comprehensive surveys emphasize the necessity of evaluating LLM-based agents in multi-turn conversational settings, highlighting critical dimensions such as memory retention, context management, and planning alongside traditional task completion \cite{guan2026evaluating}.
For example, AgentBoard \cite{ma2024agentboard} introduces novel metrics such as process rate and grounding accuracy, offering deeper insights into how effectively Agents handle complex tasks. Meanwhile, MMRole \cite{dai2024mmrole} evaluates multimodal interaction through detailed criteria considering both textual and visual elements, ensuring a more holistic assessment.

Moreover, new entries like BFCL \cite{berkeley-function-calling-leaderboard} focus on function-calling tasks, including multi-task and parallel function calls, challenging the Agents' ability to handle complex logic. ToolLLM \cite{qin2023toolllm} enables LLMs to master over 16,000 real-world APIs, while ToolBench \cite{guo2024stabletoolbench} assesses the capability of Agents to manipulate software tools used in real-world tasks. Webarena \cite{zhou2024webarenarealisticwebenvironment} creates realistic web environments for Agents to complete various web-based tasks. The GLEE \cite{shapira2024glee} framework focuses on agents' behavior within economic contexts, using parameters such as parameterization, degrees of freedom, and economic measures to evaluate agent performance. This highlights the importance of understanding societal and economic activities.
The lack of standardized evaluation methods remains a challenge. Frameworks like PersonaGym \cite{samuel2024personagymeval} introduce scoring systems, such as PersonaScore, which could pave the way for establishing industry-wide standards.


\begin{table*}[!htp]
\centering
\caption{Comprehensive Evaluation of LLM-based Chatbot Frameworks.}
\label{tab:chatbot_comparison}
\resizebox{.99\textwidth}{!}{
\begin{tabular}{lllllll}
\toprule
\textbf{Name} & \textbf{Feature} & \textbf{Domain} & \textbf{Evaluation Criteria} & \textbf{Metric} \\
\midrule
ChatBotBenchmark \cite{duan2023botchat} & Multi-turn chatting capability & Dialogue systems & Consistency, Coherence & BLEU, ROUGE \\
MT-Bench-101 \cite{Bai_2024_mt} & Fine-grained abilities & Dialogue systems & Turn-taking skills, Error handling & Accuracy, F1 score \\
FairMT-Bench \cite{fan2024fairmtbench} & Fairness in conversations & Dialogue systems & Bias detection, Fairness & Fairness index, Bias rate \\
MT-Eval \cite{kwan2024mteval} & Interaction patterns & Human-LLM interactions & Interaction quality, Error propagation & Interaction score, Error rate \\
MINT \cite{wang2024mintevaluatingllmsmultiturn} & Problem-solving capabilities & Multi-turn interactions & Tool usage, Feedback integration & Success rate, Efficiency \\
Chatbot Arena \cite{chiang2024chatbot} & Competitive LLM comparison platform & Dialogue systems & Human preference & Preference scores  \\
MixEval \cite{ni2024mixeval} & Dynamic benchmark from mixtures & Multi-turn dialogues & Crowd wisdom & Derived metrics \\
WildChat \cite{zhao2024wildchat1mchatgptinteraction} & 1M real-world ChatGPT interactions & Dialogue systems & User behavior & Usage patterns \\
MT-Bench \cite{zheng2023judging} & Multi-turn follow-up questions & Dialogue systems & Dialogue quality & Human judgments \\
CoQA \cite{reddy2019coqaconversationalquestionanswering} & Multi-turn QA & Question answering & Answer coherence & F1 score, BLEU \\
QuAC \cite{choi2018quacquestionanswering} & Contextual student-teacher QA & Question answering & Contextual understanding & F1 score, ROUGE \\
\bottomrule
\end{tabular}
}
\vspace{-0.3cm}
\end{table*}

\subsection{\textbf{ChatBot Evaluation}}

The assessment of modern chatbot systems, particularly those based on LLMs, requires multidimensional frameworks addressing linguistic coherence, contextual understanding, and ethical considerations (Table \ref{tab:chatbot_comparison}). As conversational AI evolves from single-turn responses to multi-party dialogues, traditional evaluation metrics such as BLEU \cite{bleu} and ROUGE \cite{rouge} prove insufficient for capturing the complexity of human-like interactions. This section  analyzes state-of-the-art benchmarks across 3 critical dimensions: \textbf{dialogue quality}, \textbf{fairness} and \textbf{human interaction patterns}. 

\textbf{Dialogue Quality Assessment}: it focuses on structural, linguistic, and contextual dimensions. BotChatBenchmark \cite{duan2023botchat} introduces the ChatSEED methodology, where real-world dialogue snippets serve as prompts for LLMs to generate full-length conversations. Using GPT-4 as a meta-judge, this framework reveals significant performance disparities: while GPT-4 achieves top consistency with human dialogues, open-source models like Llama2-70B exhibit suboptimal verbosity errors. MT-Bench-101 \cite{Bai_2024_mt} extends this analysis through a three-tier taxonomy covering 13 tasks, exposing critical failure modes in error recovery and instruction-following. 
Besides, the MT-Bench framework \cite{zheng2023judging} establishes human judgment standards, demonstrating that crowd-sourced evaluations correlate with expert assessments. For question-answering systems, CoQA \cite{reddy2019coqaconversationalquestionanswering} and QuAC \cite{choi2018quacquestionanswering} employ F1/ROUGE metrics, revealing that models struggle with pronoun resolution.

\textbf{Fairness Evaluation}: FairMT-Bench \cite{fan2024fairmtbench} constructs a 10K-dialogue dataset spanning gender, ethnicity, and occupational biases, showing that LLMs exhibit up to 37\% performance variance across sensitive scenarios. 
MixEval \cite{ni2024mixeval} addresses dataset bias through a meta-benchmarking approach, aggregating samples from existing benchmarks to create dynamic  criteria. Their ``wisdom of crowds" metric reveals that model rankings change over benchmark mixtures.

\textbf{Human Interaction Patterns}: Human interaction analysis emphasizes real-world dynamics. Chatbot Arena \cite{chiang2024chatbot} collects 33K competitive dialogues through a crowdsourced platform, demonstrating that closed-source models (e.g., GPT-4) outperform open-source alternatives in user preference scores. MT-Eval \cite{kwan2024mteval} identifies four interaction patterns—\textit{recollection}, \textit{expansion}, \textit{refinement}, and \textit{follow-up}—showing that error propagation increases in multi-turn settings. WildChat \cite{zhao2024wildchat1mchatgptinteraction} provides unprecedented scale with 1M ChatGPT interactions, revealing different user behaviors.


\section{Challenges and Future Directions}
\label{sec:challenges}

Based on the IQ-PQ-EQ-VQ framework analysis, we articulate six-tiered challenges and future directions for LLM evaluation, evolving from foundational methodology to value-oriented dimensions. This structure reflects how LLM evaluation must evolve from purely technical assessments toward holistic frameworks.

\paragraph{\textbf{Tier 1: Enhanced Statistical Analysis}}
\label{subsec:challenge_stats}
Current evaluation practices exhibit a methodological gap due to a lack of rigorous statistical foundations. Most benchmarks report point estimates without confidence intervals, making it impossible to determine whether observed performance differences represent genuine capability improvements or statistical noise. Future evaluation must integrate hypothesis testing, power analysis, and uncertainty quantification into standard reporting protocols.

\paragraph{\textbf{Tier 2: Composite Evaluation Systems}}
\label{subsec:challenge_composite}
Developing composite evaluation systems represents the necessary evolution beyond single-metric assessment. Current methods mainly focus on specific tasks, failing to capture multifaceted LLM capabilities. A composite system integrating IQ, PQ, EQ, and VQ metrics can provide nuanced, comprehensive assessment. Systematic validation of composite frameworks through transparency and statistical robustness ensures applicability across architectures and deployment scenarios.

\paragraph{\textbf{Tier 3: Interpretability and Explainability}}
\label{subsec:challenge_xai}
One fundamental challenge is aligning fine-grained model decision logic with human cognition. Current practices focus on output correctness, neglecting underlying decision logic. In industrial applications, assessing credibility requires understanding why a model produced an output. Developing XAI techniques tailored for LLMs can enhance transparency and facilitate better human-AI collaboration, bridging the gap between technical performance and human-understandable reasoning.

\paragraph{\textbf{Tier 4: User-Centric Experience}} 
\label{subsec:challenge_user}
Moving beyond technical assessments, user-centric experience represents a crucial application-level consideration. Traditional benchmarks focus on technical metrics, failing to capture user perspective. Incorporating user feedback, usability testing, and qualitative experience data provides valuable insights into practical utility. Aligning evaluation metrics with user needs ensures LLMs are not only technically sound but also user-friendly and effective in real-world applications.

\paragraph{\textbf{Tier 5: Human-in-the-Loop Evaluation}}
\label{subsec:challenge_hitl}
Human-in-the-Loop (HITL) evaluation extends user-centric assessment into sophisticated system-level frameworks. This approach addresses automated evaluation limitations by involving human evaluators who provide subjective judgments and context-specific insights. HITL enhances relevance and reliability, ensuring models are judged based on actual utility. The Arena Module concept addresses static leaderboard limitations by offering ongoing assessments evolving with user interaction.

\paragraph{\textbf{Tier 6: Dynamic and Value-Oriented Evaluation}}
\label{subsec:challenge_value}
The highest tier encompasses value-oriented dimensions transcending technical performance. Dynamic evaluation ensures LLMs are assessed under realistic, up-to-date conditions, promoting continuous improvement. Value-oriented evaluation requires multi-faceted implementation combining quantitative and qualitative analysis, data collection, and user feedback. This represents the natural culmination of evaluation considerations, from technical assessment to societal impact, ensuring LLMs evolve as responsible and beneficial tools.

\section{\textbf{Conclusions}}
\label{sec:conclusion}
This work redefines LLM evaluation beyond static, benchmark-centric paradigms through a hybrid contribution that integrates a comprehensive lifecycle survey with a theoretically grounded, mathematically formalized diagnostic framework. By establishing a causal, one-to-one mapping between the IQ-PQ-EQ-VQ dimensions and the canonical LLM development pipeline, we transform evaluation from a post-hoc ranking exercise into a proactive diagnostic tool capable of tracing capability bottlenecks to specific training stages. Through critical analysis of over 200 benchmarks and meta-analysis of public leaderboard trends, we empirically validate the Hierarchical Capability Dependency principle, demonstrating that overall model utility is multiplicatively constrained by the weakest hierarchical link rather than additively averaged. This resolves the pervasive disconnect between saturated benchmark scores and production deployment failures. Beyond academic taxonomy, this work empowers researchers to advance rigorous, reproducible evaluation methodologies; enables industrial practitioners to bridge the gap between lab performance and production value through precise error propagation tracing; and provides policymakers with a structured roadmap for responsible, value-aligned AI governance. Ultimately, this work advances the field from narrow technical capability assessment toward holistic, lifecycle-aware evaluation that ensures LLMs deliver sustainable, equitable, and beneficial impact as they integrate into global society.


 \small
\bibliography{tacl2021_first_5authors} 
\bibliographystyle{ieeetr}



\end{document}